\documentclass[journal]{IEEEtai}

\usepackage[colorlinks,urlcolor=blue,linkcolor=blue,citecolor=blue]{hyperref}

\usepackage{color,array}

\usepackage{graphicx}
\usepackage{times}
\usepackage{epsfig}
\usepackage{graphicx}
\usepackage{amsmath}
\usepackage{amssymb}
\usepackage{algorithm}
\usepackage{algpseudocode}
\usepackage{amsmath}
\usepackage{multirow}
\usepackage{subfigure}
\usepackage{booktabs}
\usepackage{color}
\usepackage{booktabs}
\usepackage{hyperref}
\usepackage{dutchcal}

\definecolor{revise_color}{RGB}{0, 0, 0}
\definecolor{R2}{RGB}{0, 0, 0}

\begin{document}

\title{Adversarial Graph Disentanglement with Component-specific Aggregation}

\author{Shuai Zheng,
        Zhenfeng Zhu,
        Zhizhe Liu,
        Jian Cheng,
        and~Yao~Zhao,~\IEEEmembership{Fellow,~IEEE}% <-this % stops a space
\thanks{This work was supported in part by Science and Technology Innovation 2030 – New Generation Artificial Intelligence Major Project under Grant No. 2018AAA0102100, National Natural Science Foundation of China under Grant No. 61976018, Beijing Natural Science Foundation under Grant No. 7222313. \textit{(Corresponding Author: Zhenfeng Zhu)}}
\thanks{S. Zheng, Z. Zhu, Z. Liu, and Y. Zhao are with the Institute of Information Science, Beijing Jiaotong University, Beijing 100044, China, and also with the Beijing Key
	Laboratory of Advanced Information Science and Network Technology,
	Beijing 100044, China (e-mail: zs1997@bjtu.edu.cn; zhfzhu@bjtu.edu.cn; zhzliu@bjtu.edu.cn;
	yzhao@bjtu.edu.cn).}
\thanks{J. Cheng is with the National Laboratory of Pattern Recognition, Institute of Automation, Chinese Academy of Sciences and University of Chinese
	Academy of Sciences, Beijing 100190, China (e-mail: jcheng@nlpr.ia.ac.cn).}
% <-this % stops a space
}

\markboth{Journal of IEEE Transactions on Artificial Intelligence, Vol. 00, No. 0, Month 2020}
{First A. Author \MakeLowercase{\textit{et al.}}: Bare Demo of IEEEtai.cls for IEEE Journals of IEEE Transactions on Artificial Intelligence}

\maketitle

\begin{abstract}
A real-world graph has a complex topological structure, which is often formed by the interaction of different latent factors.
%
%Factor entanglement limits the expressiveness of the node representation, making oversmoothing prone to occur in model training
Disentanglement of these latent factors can effectively improve the robustness and expressiveness of the node representation of a graph.
However, most existing methods lack consideration of the intrinsic differences in relations between nodes caused by factor entanglement. 
In this paper, we propose an \underline{\textbf{A}}dversarial \underline{\textbf{D}}isentangled \underline{\textbf{G}}raph \underline{\textbf{C}}onvolutional \underline{\textbf{N}}etwork (ADGCN) for disentangled graph representation learning. To begin with, we point out two aspects of graph disentanglement that need to be considered, i.e., micro-disentanglement and macro-disentanglement.
For them, a component-specific aggregation approach is proposed to achieve micro-disentanglement by inferring latent components that caused the links between nodes.
On the basis of micro-disentanglement, we further propose a macro-disentanglement adversarial regularizer to improve the separability among component distributions, thus restricting the interdependence among components.
%
%Additionally, to learn collaboratively a better disentangled representation and topological structure,
%a diversity-preserving node sampling based progressive refinement of graph structure is proposed.
Additionally, to reveal the topological graph structure, a diversity-preserving node sampling approach is proposed, by which the graph structure can be progressively refined in a way of local structure awareness.
The experimental results on various real-world graph data verify that our ADGCN obtains more favorable performance over currently available alternatives. The source codes of ADGCN are available at \textit{\url{https://github.com/SsGood/ADGCN}}.
\end{abstract}

\begin{IEEEImpStatement}
Graph representation learning aims to represent discrete irregular graph data as continuous vectors. However, most of the existing methods stay on the general utilization of the external topological structure to learn representation, ignoring the analysis of the internal causes of the graph. To tackle these concerns, disentangled graph representation learning has been seen as an effective solution. In this paper, we specify two aspects of graph disentanglement that need to be considered and propose an adversarial disentangled graph representation learning method. Our extensive experiments show that the proposed method not only achieves comparable performance to state-of-the-art methods for various node classification tasks but also demonstrates some significant graph disentanglement properties. Owning to the effective graph disentanglement, our model also shows better defensive ability than previous graph neural networks. As a whole, the idea presented in this work provides a new way towards graph disentanglement.
\end{IEEEImpStatement}


\section{Introduction}

\IEEEPARstart{T}{he} graph is a general description of data and their relations, which can model a variety of scenarios, from biomedical networks to citation networks, or other things defined by relations.
There are a variety of graph-based applications in the machine learning field, such as link prediction~\cite{SDNE}, node classification~\cite{GAE, social}, and recommendation~\cite{deepwalk, node2vec}.
As an important part of graph analysis, graph representation learning is the basis for graph data to be processed by conventional machine learning methods.
%Recently, a great deal of work has promoted the development of graph representation learning.

Most graph representation learning methods can be broadly classified into two categories: proximity-based methods and neural network-based methods.
Proximity-based methods, such as DeepWalk~\cite{deepwalk}, node2vec~\cite{node2vec}, and LINE~\cite{LINE}, focus on extracting the patterns from the graph data.
These methods tend to learn the node representation with various order proximity preserving by constructing the node paths through random walks or local neighborhoods. Besides,~\cite{qiu2018network} has demonstrated that matrix factorization methods~\cite{grarep} are also equivalent to proximity-based approaches.
While neural network-based methods usually apply graph convolution to aggregate neighborhood information~\cite{GCN, GAT, Graphsage}, or use autoencoder to merge structure and node features~\cite{SDNE, GAE}.

\begin{figure}[t]	
	\centering
	\includegraphics[width=3.5in]{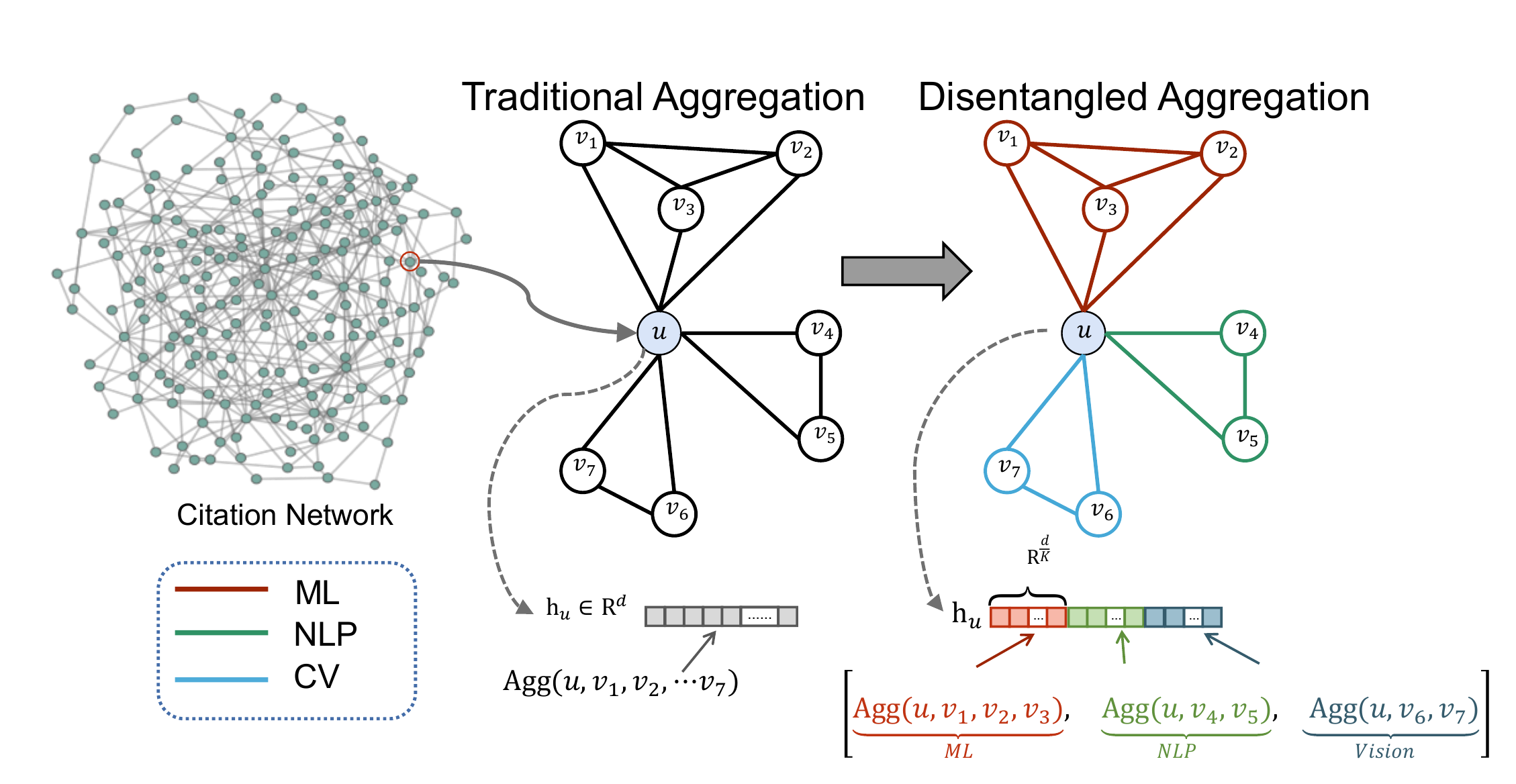}
 \vspace{-0.5cm}
	\caption{A simple example in a citation network. A paper $u$ for visual question answering (VQA) may simultaneously have citation relationships with papers in the fields of ML, NLP, and CV because of the related researches it involves. Through differentiated aggregation, the learned separable representation by graph disentanglement can reveal more about the latent semantics among different neighborhoods.
	}
	\label{example}
\vspace{-0.5cm}	
\end{figure}

Significant progress in graph representation learning has been made recently, especially the emergence of graph convolutional networks (GCNs)~\cite{survey}. Nevertheless, most of the existing methods learn node representation by treating the local neighborhood or node path as a whole, ignoring the subtle differences between node neighbors. Actually, such practices make the model prone to learning over-smoothed representations when a network goes deeper~\cite{powerful}.
In general, a real-world graph is relatively complex and formed by the interaction of a variety of latent factors.
In other words, links between a node and its neighbors may be caused by different components, which is why nuances are presented between node neighbors. However, these nuances are often overlooked by existing methods, limiting the expressiveness of learned node representations. Instead, we would like to consider the distinction of node neighbors and disentangle the node representation towards discovering those distinct latent components with richer information, i.e., graph disentanglement.

As an example in Fig.~\ref{example}, unlike traditional undifferentiated aggregation of neighboring nodes by an aggregation function, graph disentanglement aims to learn the node representation that consists of some distinct latent components with each containing a certain semantic.
Thus, how to disentangle the node representation into multiple components with distinct latent semantics is of great significance for graph representation learning, which will also be helpful for improving the robustness and interpretability of graph neural networks.

In the field of image analysis and understanding, disentangled representation learning has been successfully applied to a variety of tasks, such as foreground separation and visual reasoning~\cite{van2019disentangled}. 
\textcolor{black}{It aims to separate the distinct, informative factors of variations in the data and learn component representations corresponding to the factors accordingly~\cite{definition_dis}.}
However, less focus has been put on disentangled representation learning of graph~\cite{DisenGCN, IPGDN, FactorGCN}. 
\textcolor{black}{According to the above notion of disentanglement~\cite{definition_dis}, there are two aspects to consider to implement node-level graph disentanglement. 1) For a given graph, we need to infer the informative component representation of each node in the graph corresponding to the different factors. 2) Besides, we need to ensure the component subspaces are as separable as possible. The former can be understood as representation-level \textit{\textbf{micro-disentanglement}} and the latter can be understood as subspace-level \textit{\textbf{macro-disentanglement}}.}
%
%DisenGCN~\cite{DisenGCN} is the first work that attempts to achieve graph disentanglement and made a preliminary demonstration.
%
% Different from those holistic methods like GCN~\cite{GCN}, ~\cite{DisenGCN} and~\cite{IPGDN} have put the focus on finding the difference of latent semantics among node neighborhood information and utilizing a neighborhood routing mechanism to infer the latent factors by iteratively analyzing the clusters of component spaces formed by nodes and their neighbors, thereby learning the disentangled components of a node as representation.
%

\noindent\textcolor{R2}{\textbf{Challenges of Graph Disentanglement:}}
Despite the essential explorations that have been made by ~\cite{DisenGCN} and~\cite{IPGDN}, there are still some key issues that should be addressed:
\textbf{(i)} The existing methods attempt to perform micro-disentanglement through neighborhood routing mechanisms.
% by projecting node's neighborhoods into specific multiple component spaces.
Still, this practice lacks consideration that the relationship between nodes may be formed by a combination of multiple factors. %Inevitably, it will cause the aggregated representation to be none representative of the latent semantics associated with the component;
\textbf{(ii)} In addition to the micro-disentanglement, the components should also be macro-separable at the component distribution level, while DisenGCN neglects to restrict interdependence among component distributions; \textbf{(iii)} The disentangled representations should be inversely further reveal the intrinsic graph structure based on the given initial adjacency matrix.

\textcolor{revise_color}{To address the above key issues, we propose a novel \underline{A}dversarial \underline{D}isentangled \underline{G}raph \underline{C}onvolutional \underline{N}etwork (ADGCN), for disentangled graph representation learning.}
Specifically, different from DisenGCN and IPGDN, 
\textcolor{R2}{to address the micro-disentanglement of a combination of multiple factors,}
we 
% propose a dynamic assignment mechanism by allowing for the existence of multiple relationships between nodes. 
consider the intra-component correlation and inter-component separability simultaneously, and achieve component-specific aggregation for the micro-disentanglement of the node.
\textcolor{R2}{On the basis of the micro-disentanglement performed by the component-specific aggregation, 
conditional adversarial learning is further deployed to restrict the interdependence among components for the macro-disentanglement.} To the best of our knowledge, this is the first attempt to introduce adversarial learning for graph disentanglement.
%
% \textcolor{R2}{Since the disentangled node representations, in turn, can be utilized to uncover latent node associations, thereby providing a more fine-grained graph topology,}
 The learned disentangled representation is also leveraged to explore the underlying structure of the graph in a progressive manner.
we propose a diversity-preserving node sampling method to guide the local structure-aware graph refinement and spread to the overall graph further. In summary, the main contributions can be highlighted as follows:
\begin{itemize}
	% \item 
 % We propose a novel \underline{A}dversarial \underline{D}isentangled \underline{G}raph \underline{C}onvolutional \underline{N}etwork (ADGCN), for disentangled graph representation learning, thus to unveil the associated multiple latent components.
	\item
	To achieve micro-disentanglement for each node, a component-specific aggregation with a dynamic assignment mechanism is proposed, which ensures the compactness of the component space and allocates the neighbors of the target node to the corresponding component space to realize micro-disentanglement.
	\item
	For implementing macro-disentanglement, we propose a conditional adversarial regularizer to improve the separability among component distributions, thus learning the more distinct components.
	%to learn the latent components with weak correlation.
	\item
	To refine the global graph structure in a progressive way, we propose a diversity-preserving node sampling approach to perform local structure-aware refinement. Hence, the latent global topological graph structure can be well revealed to benefit graph disentanglement.
\end{itemize}

%%%%%%%%%%%%%%%%%%%%%%%%%%%%%%%%%%%%%%%%%%%%%%%%%%%
\begin{figure*}[t]	
	\centering
	\includegraphics[width=5.5in]{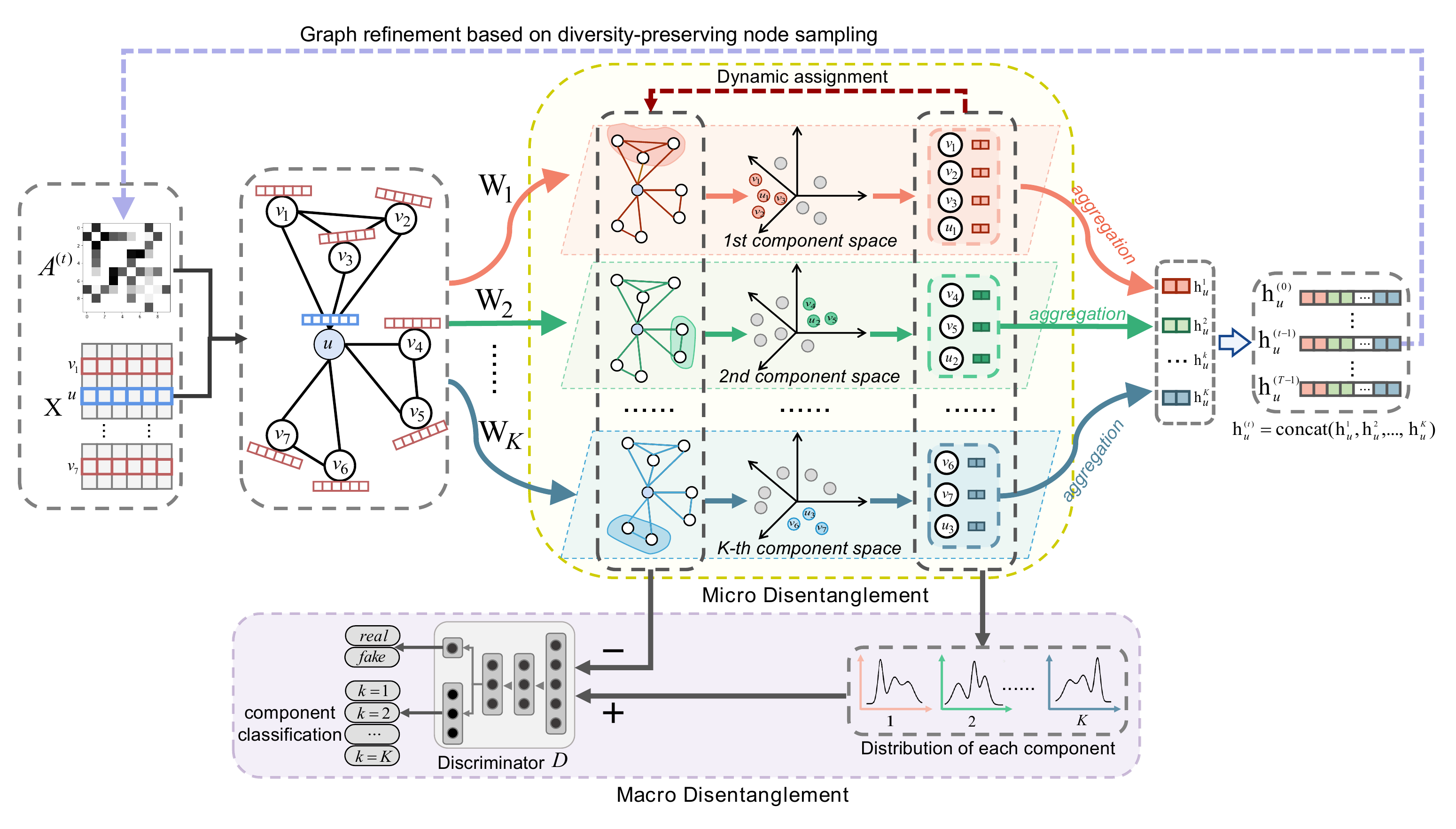}
        \vspace{-10pt}
	\caption{Architecture overview of a single-layer ADGCN. $A^{(t)}$ represent the refined adjacency at epoch $t$.
		$\mathbf{X}$ and $\mathbf{W}_{k}$ denote the node raw feature and the projection matrix to $k$-th component space respectively.
		$\{\mathbf{h}^{k}_{u}\}^{K}_{k=1}$ represent the components of node $u$ and $\mathbf{h}^{(t)}_{u}$ as the output denotes the disentangled representations at epoch $t$, where $\mathbf{h}^{(t)}_{u}$ is used to update $A^{(t)}$ to $A^{(t+1)}$ through sample-based graph refinement.
		% '$+$' and '$-$' represent the real and fake samples that are sampled from $\mathbf{h}^{(t)}_{u}$ and $\mathbf{c}^{(t)}_{u}$, respectively.
  }
	\label{ADGCN}
\vspace{-0.5cm}
\end{figure*}

\section{Related Work}
\subsection{Graph neural networks for graph disentanglement}

Currently, several studies on disentangled representation learning of graph data have emerged. DisenGCN~\cite{DisenGCN} first studied the problem of graph disentanglement at the node level. On this basis, IPGDN~\cite{IPGDN} enhanced the independence constraint of graph disentanglement by using HSIC. Besides, FactorGCN~\cite{FactorGCN} proposed graph-level disentangling for graph classification. SGCN~\cite{SGCN} combines \cite{FactorGCN} and \cite{DisenGCN} in a unified framework. Recently, \cite{LGD-GCN} proposes to leverage neighborhood routing locally and message passing globally to improve the graph disentanglement. While~\cite{DisGNN} considers hierarchical edge disentanglement to assist node classification tasks. Meanwhile, DSSL~\cite{DSSL} and DGCL~\cite{DGCL} consider graph disentanglement in self-supervised learning and contrastive learning settings, respectively.
In this paper, according to the concept of disentanglement~\cite{definition_dis}, we give two aspects that need to be considered for node-level graph disentanglement. Furthermore, the proposed ADGCN is the first work that introduces adversarial learning for graph disentanglement. It not only considers the disentanglement of individual nodes but also imposes constraints at the component distribution level.

% In contrast to \cite{DisenGCN} and \cite{IPGDN}, ADGCN not only considers the disentanglement of individual nodes but also imposes constraints at the component distribution level.

\subsection{Adversarial learning on graphs}
As a powerful generative model, generative adversarial networks have been proposed by Goodfellow et al.~\cite{GAN}. Then, ~\cite{molgan} and~\cite{netgan} successfully applied adversarial learning to handle graph data for graph generation tasks.
Besides, the success of~\cite{BiGAN,ALI} shows that GAN and adversarial learning also have excellent performance for data understanding. Benefiting from the great ability of distribution fitting, adversarial learning has also been adopted to the domain adaption field~\cite{DA_survey} as a critical part.
Hence, some studies have also applied adversarial learning to the field of graph representation learning.
\cite{graphgan} generated node pairs to approximate the linkage distribution, enhancing the structure-preserving capability of the node representation. And \cite{asine} also proposed to sample sub-graph to learn node embedding based on adversarial learning. Another common practice, as in ~\cite{ARGA} and ~\cite{DBGAN}, is to utilize adversarial learning as a regularizer to constrain the representation distribution of nodes, thereby improving the generalization ability of the representations. 
% AIDW~\cite{AIDW} adopted a similar form to introduce adversarial learning to~\cite{deepwalk}.
In addition, DGI~\cite{DGI} constructed degraded graphs for adversarial learning with the original graph.
~\cite{lu2020} maintained consistency of node representation between multiple graphs through adversarial learning.
Unlike them, we apply an adversarial regularizer to control the separability between component distributions for graph disentanglement.

\section{Methodology}

%In this section, we first introduce some notations and the overall framework of ADGCN, then present the component-specific aggregation for micro-disentanglement and an adversarial regularizer for macro-disentanglement, finally introduce a node sampling based progressive graph refinement strategy.

\subsection{Problem Formulation and Overview of Framework}

To begin with some definitions, an undirected graph is given as $\mathcal{G} = (V,E)$, where $V$ consists of the set of nodes with $|V| = n$, and $E$ is the set of edges among nodes. To represent the link strength (similarity) among graph nodes, we use $\mathbf{A}\in R^{n\times n}$ to denote the adjacency matrix\footnote{For a graph with the hard link, we have $\mathbf{A}_{i,j}$ = 1 if there exists an edge $(i,j) \in E$, otherwise $\mathbf{A}_{i,j}= 0$.}, and  $\mathbf{D}=diag([\mathbf{d}_{u}]_{u=1,\cdots,n})$ to denote the corresponding degree matrix with $d_{u}=\sum_{i=1}^{n}\mathbf{A}_{u,i}$. Meanwhile, let $\mathbf{X}=[\mathbf{x}_{1},\cdots,\mathbf{x}_{n}]\in\mathbb{R}^{n \times f}$ denote the node feature matrix with $\mathbf{x}_{u}\in\mathbb{R}^{f}$, $u=1,\cdots,n$, representing the raw feature of  the $u$-th node, and $\mathbf{H}=[\mathbf{h}_{1},\cdots,\mathbf{h}_{n}]\in\mathbb{R}^{n \times d}$ the node representation matrix.
In addition,
$\mathbf{N}(u)$ is defined as a node subset containing neighbors of node $u$ and $\tilde{\mathbf{N}}(u) = \mathbf{N}(u) \cup \{u\}$.

\noindent
\textit{1) Micro-Disentanglement:} Assuming that $\mathcal{G}$ is formed by $K$ latent factors corresponding to $K$ component spaces. For \underline{a specific node $u$}, we want to decompose its latent representation $\mathbf{h}_{u} $ into $K$ component representations $\{\mathbf{h}^{1}_{u}, \mathbf{h}^{2}_{u}, \cdots, \mathbf{h}^{K}_{u}\}$, where $\mathbf{h}^{k}_{u} \in \mathbb{R}^{\frac{d}{K}}$ corresponds to the $k$-th factor.
%For a specific node $u$, we expect that its component representation $\mathbf{h}^{k}_{u}$ only sensitive to the variation in the $k$-th factor, while not affected by the one in other factors~\cite{RDCM}.

\noindent
\textit{2) Macro-Disentanglement:}
On the basis of micro-disentanglement, for \underline{all nodes $V$} in the given graph, we still expect the component representations lying on different component spaces to be far away from each other, i.e., the difference between any two component distributions should be evident.

\noindent
\textcolor{R2}{\textit{3) Graph Refinement:}
After the disentanglement mentioned above, the disentangled representations enable a more fine-grained exploration of node associations, which, allows for a further revelation of the intrinsic structure of the graph based on the given initial adjacency matrix.}

\noindent
\textit{4) Overview of Framework:} The overall illustration of the proposed ADGCN model is shown in Fig.~\ref{ADGCN}.
\textcolor{R2}{
For a specific node $u$ and its neighbors in the given graph, the initial embeddings $\{\mathbf{c}^{k}_{i}\}^{K}_{k=1}$ are first obtained from $\mathbf{x}_{i}$ through projection matrices $\mathbf{W} = \{\mathbf{W}_{1}, \mathbf{W}_{2}, \cdots, \mathbf{W}_{K}\}$, where $i \in \tilde{\mathbf{N}}(u)$.
\\
\textbf{To perform the micro-disentanglement}, $\{\mathbf{c}^{k}_{i}\}_{i \in \tilde{\mathbf{N}}(u)}$ are aggregated into $\mathbf{h}^{k}_{u}$ in the $k$-th component space through the dynamic assignment mechanism and component-specific aggregation, then we have $\mathbf{h}_{u}=\textrm{concat}(\mathbf{h}^{1}_{u},\mathbf{h}^{2}_{u},\cdots,\mathbf{h}^{K}_{u})$.
\\
\textbf{To perform the macro-disentanglement}, $\mathbf{c}^{k}_{u}$ and $\mathbf{h}^{k}_{u}$ are input into the discriminator $D$, where $\mathbf{c}^{k}_{u}$ and $\mathbf{h}^{k}_{u}$ are regarded as a \textit{"fake"} sample and a \textit{"real"} sample respectively.
\\
\textbf{For graph refinement}, if node $u$ would be selected by diversity-preserving node sampling, $\mathbf{h}_{u}$ is also used to update $\mathbf{A}^{(t-1)}$ to $\mathbf{A}^{(t)}$.
}
%
% For a specific node $u$ and its neighbors in the given graph, the initial embeddings $\{\mathbf{c}^{k}_{i}\}^{K}_{k=1}$ are obtained from $\mathbf{x}_{i}$ through projection matrices $\mathbf{W} = \{\mathbf{W}_{1}, \mathbf{W}_{2}, \cdots, \mathbf{W}_{K}\}$, where $i \in \tilde{\mathbf{N}}(u)$.
%
% Then, through the component-specific aggregation, $\{\mathbf{c}^{k}_{i}\}_{i \in \tilde{\mathbf{N}}(u)}$ are aggregated into $\mathbf{h}^{k}_{u}$ in the $k$-th component space, and the concatenation of $\{\mathbf{h}^{k}_{u}\}^{K}_{k=1}$ is $\mathbf{h}_{u}=\textrm{concat}(\mathbf{h}^{1}_{u},\mathbf{h}^{2}_{u},\cdots,\mathbf{h}^{K}_{u})$.
%
% For adversarial learning, $\mathbf{c}^{k}_{u}$ and $\mathbf{h}^{k}_{u}$ are input into the discriminator $D$ where $\mathbf{c}^{k}_{u}$ and $\mathbf{h}^{k}_{u}$ are regarded as a \textit{"fake"} sample and a \textit{"real"} sample respectively. In addition, $\mathbf{h}_{u}$ is used for graph refining to update $\mathbf{A}^{(t-1)}$ to $\mathbf{A}^{(t)}$.

\vspace{-0.2cm}
\subsection{Component-Specific Aggregation for Micro-Disentanglement}
\label{MDCL}

\begin{figure}[t]	
	\centering
	\includegraphics[width=3in]{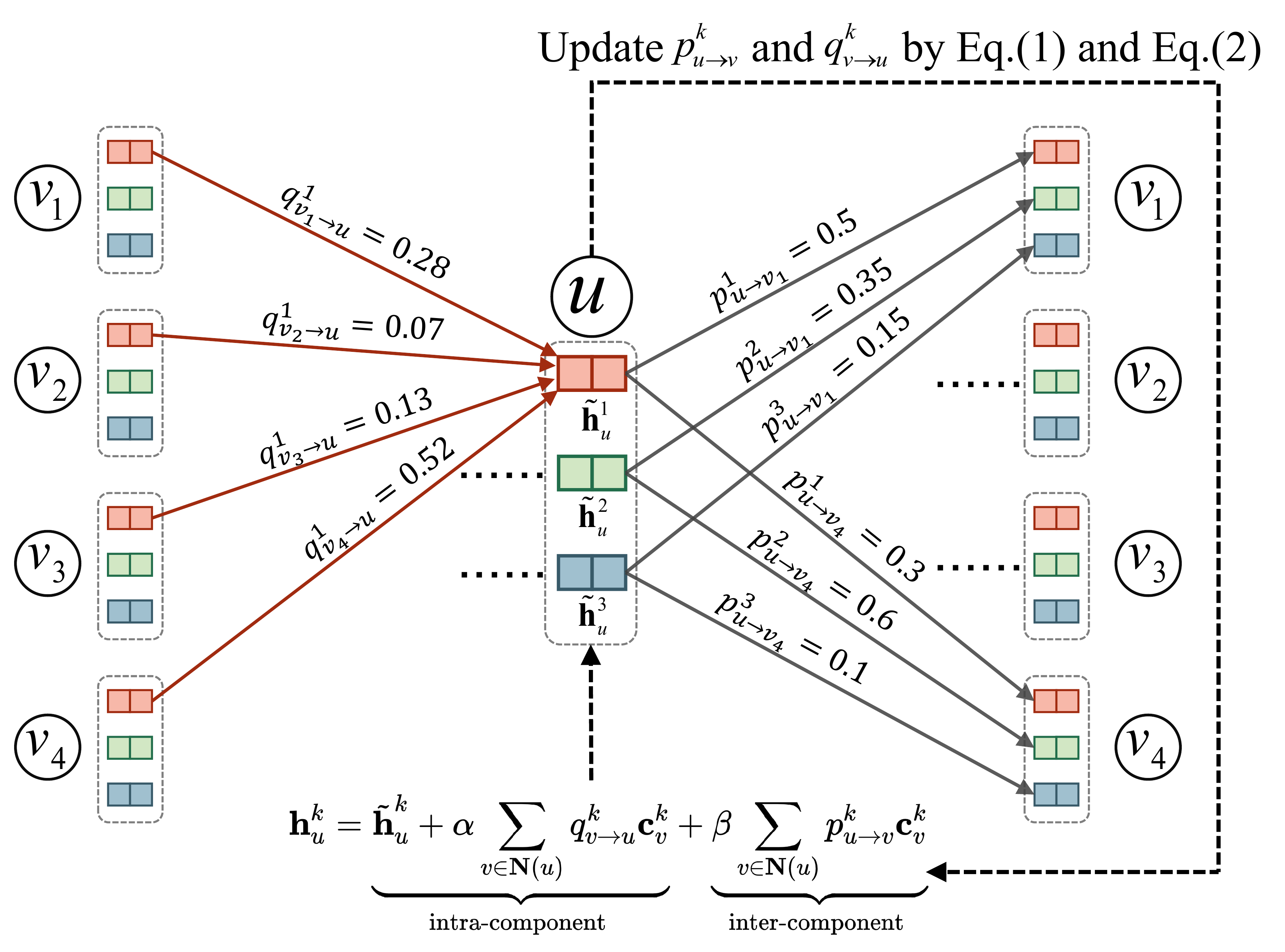}
	\caption{Illustration of component-specific aggregation based on dynamic assignment.
		The proposed dynamic assignment iteratively allocates the neighbor nodes to different component spaces according to the hubness score $p^{k}_{u \to v}$. Meanwhile, according to the authority score $q^{k}_{v \to u}$ of node $v$ in each component space, $\{\mathbf{c}^{k}_{i}\}_{i \in \tilde{\mathbf{N}}(u)}$ are aggregated into $\tilde{\mathbf{h}}^{k}_{u}$. Here, only three component spaces are considered.}
	\label{micro_disen}
	\vspace{-0.5cm}
\end{figure}
Based on an important assumption, i.e., the links between nodes are formed by the interaction of different latent factors, our ADGCN learns disentangled components by aggregating the different neighborhood information of the node to achieve the micro-disentanglement.

Intuitively, if node $u$ and its neighbor node $v$ are similar in the $k$-th component space, the link between the two nodes is likely caused by the factor $k$.
Furthermore, for a subset $\mathbf{N}_{k}(u) \subseteq \mathbf{N}(u)$, if all nodes in $\mathbf{N}_{k}(u)$ are connected to node $u$ due to factor $k$, these nodes should also be similar and form a cluster with $u$ in the $k$-th component space.
It is worth noting that the above two points correspond to the first-order and the second-order proximities in the topological structure formed by the factor $k$, respectively~\cite{LINE}. Based on these two points, DisenGCN~\cite{DisenGCN} proposed a neighborhood routing mechanism, which was also followed by~\cite{IPGDN}. However, the neighborhood routing mechanism merely considers the restriction on the separability of inter-components. It also constrains that each neighbor should belong to only one component space, ignoring multi-relations among nodes.

Motivated by the observations mentioned above, we propose a dynamic assignment mechanism as shown in Fig.~\ref{micro_disen} to achieve component-specific aggregation for micro disentanglement.
%
%Inspired by HITS~\cite{HITS}, the dynamic assignment mechanism is performed by iteratively calculating the \textbf{\textit{hubness score}} and the \textbf{\textit{authority score}} of nodes.
%
To characterize the interactive relationship between node $u$ and its neighbor node $v\in \mathbf{N}(u)$ in each component space, we define the \textbf{\textit{hubness score}} $p^{k}_{u \to v}$ and the \textbf{\textit{authority score}} $q^{k}_{v\to u}$ of node $u$ and $v$ in the $k$-th component space:
%Particularly, to characterize the interactive relationship between node $u$ and its neighbor node $v\in \mathbf{N}(u)$ in each component space, we follow the idea of HITS~\cite{HITS} by defining the \textbf{\textit{hubness score}} $p^{k}_{v}(u)$ and the \textbf{\textit{authority score}} $q^{k}_{v}(u)$ of node $v$ in the $k$-th component space as follows:
%
%where $p^{k}_{v}$ denote the probability that $v$ is connected to $u$ by factor $k$, and $q^{k}_{v}$ denote the importance of $v$ to $u$ in the $k$-th component space.
%Then, we define the \textbf{\textit{hub score}} of $v$ in the $k$-th component space as $p^{k}_{v}$, which denote the probability that $v$ is connected to $u$ by factor $k$. And we define the \textbf{\textit{authority score}} of $v$ in the $k$-th component space as $q^{k}_{v}$, which denote the importance of $v$ to $u$ in the $k$-th component space.
\begin{equation}
p^{k}_{u \to v} = \frac{\mathrm{exp}(s^{k}_{uv})}{\sum_{k=1}^{K}\mathrm{exp}(s^{k}_{uv})},\ \  q^{k}_{v\to u} = \frac{\mathrm{exp}(s^{k}_{uv})}{\sum\limits_{v\in \mathbf{N}(u)}\mathrm{exp}(s^{k}_{uv})}
\label{p}
\vspace{-0.2cm}
\end{equation}
% \begin{equation}
% q^{k}_{v\to u} = \frac{\mathrm{exp}(s^{k}_{uv})}{\sum\limits_{v\in \mathbf{N}(u)}\mathrm{exp}(s^{k}_{uv})}
% \label{q}
% \end{equation}
where  $s^{k}_{uv} = \cos(\mathbf{h}^{k}_{u}, \mathbf{c}^{k}_{v})$ denotes the similarity between the target node $u$ and its neighbor $v$ in the $k$-th component space\footnote{To avoid numerical instability caused by inconsistent norms between different component spaces, $l_{2}$-normalization is adopted for $h^{k}_{u}$ and $c^{k}_{v}$ to facilitate the calculation of similarity.}. 

In fact, as a  bidirectional information interaction between node $u$ and its neighbor $v$, the hubness score $p^{k}_{u \to v}$ means the probability that $v$ is connected to $u$ by the factor $k$, and the authority score $q^{k}_{v\to u}$ refers to the aggregation importance of $v$ for $u$ in the $k$-th component space. In particular, for a specific neighbor node $v$, the higher the hubness score $p^{k}_{u \to v}$, the more effective information between $v$ and $u$ has about factor $k$, therefore $v$ will tend to be allocated to the $k$-th component space for the aggregation of $u$. Meanwhile, a high authority score $q^{k}_{v\to u}$ means a high similarity between $u$ and $v$ in the $k$-th component space.

Based on $p^{k}_{u \to v}$ and $q^{k}_{v\to u}$, the proposed dynamic assignment mechanism can be implemented in an iterative way to obtain the aggregated representation $\mathbf{h}_u^k$ by:
%Then, different from the neighborhood routing mechanism in ~\cite{DisenGCN} for micro-disentanglement of individual sample, which merely considers the restriction on separability of inter-component, we propose a dynamic assignment mechanism.
%
%In particular, the neighbor nodes are allocated to each component space according to  $p^{k}_{v}$, and then performed weighted aggregation by using $q^{k}_{v}$. Hence, $h^{k}_{u}$ can be iteratively aggregated as:
\vspace{-0.3cm}
\begin{equation}
 {\mathbf{h}^{k}_{u}}= \underbrace{\tilde{\mathbf{h}}^{k}_{u} + \alpha \sum_{v \in \mathbf{N}(u)}q^{k}_{v\to u}\mathbf{c}^{k}_{v}}_{intra-component} + \underbrace{\beta \sum_{v \in \mathbf{N}(u)}p^{k}_{u \to v}\mathbf{c}^{k}_{v}}_{inter-component}
\label{h}	
\end{equation}
where $\tilde{\mathbf{h}}^{k}_{u}$ denotes the aggregated $\mathbf{h}^{k}_{u}$ in previous iteration, \textcolor{black}{$\alpha$ and $\beta$ are trade-off coefficients. Notably, instead of treating them as hyperparameters, we set $\alpha$ and $\beta$ as two trainable parameters and scale their range to (0,1).} The intra-component part with $q^{k}_{v\to u}$ ensures that $\tilde{h^{k}_{u}}$ tends to be aggregated into the densest area in $k$-th component space, and the inter-component part with $p^{k}_{u \to v}$ constrains the separability between $\{\mathbf{h}^{k}_{u}\}^{K}_{k=1}$. Meanwhile, different from the restricted single-relation constraint in~\cite{DisenGCN,IPGDN}, the co-aggregation of intra-component and inter-component also allows for the existence of multiple relationships between nodes.

%%%%%%%%%%%%%%%%%%%%%%%%%%%%%%%%%%%%%%%%%%%%%%%%%%%%%%%%%%%%%%%%%%%%%%%%%%%%%%%%%%%%%%%%%%%%%%%%%%%%%%%%%%%%%%%%%%%%%%%%%%%%
To make the proposed  ADGCN be capable of inductive learning, we take a similar way as in~\cite{Graphsage} by randomly sampling the neighboring nodes of $u$ to get $\mathbf{N}(u)$ with a fixed size.
Notably, average sampling is used for unweighted graphs, and probability sampling based on edge weights is applied for weighted graphs.
The pseudocode of the component-specific node representation aggregation is summarized in \textbf{Alg}. \ref{algorithm1}.

%In practice, similar to GCN\cite{GCN}, we stack multiple blocks with each consisting of a basic component-specific aggregation layer to capture higher-order proximity information. Besides, the number of components at each layers can be different which implicitly learn hierarchical representations. The pseudocode of the component-specific aggregation layer is summarized in Algorithm \ref{algorithm1}.
\renewcommand{\algorithmicrequire}{\textbf{Input:}}
\renewcommand{\algorithmicensure}{\textbf{Output:}}
\begin{algorithm}[t]
	\caption{Procedure of the component-specific aggregation for Micro-disentanglement}
	\footnotesize
	\label{algorithm1}
	\begin{algorithmic}[1] 
		\Require $\left \{ \mathbf{x}_i \in \mathbb{R}^f\right \} $: the set of node feature vectors, $i \in \tilde{\mathbf{N}}\left ( u \right )$;
		%$K$: number of component
		
		%$\tilde{T}$: iterations of dynamic assignment
		
		%$\left \{ \mathbf{W}_k \right \}$: projection matrix, $k\in K$
		\Ensure $\mathbf{h}_u$, $\mathbf{c}_u$;
		\renewcommand{\algorithmicensure}{\textbf{trainable paramters:}}
		\Ensure \textcolor{black}{$\mathbf{W}$: projection matrices.}
		
		\textcolor{black}{$\alpha, \beta$: coefficients of component-specific aggregation.}
		
		\renewcommand{\algorithmicensure}{\textbf{Hyper-paramters:}}
		\Ensure $K$: number of component.
		
		$\tilde{T}$: iterations of dynamic assignment.

		\For{$i\in \tilde{\mathbf{N}}\left( u \right)$}
		\For{$k=1$ to $K$}
		
		\State $\mathbf{c}^{k}_i \leftarrow \sigma \left ( \mathbf{W}_k \mathbf{x}_i \right )$
		
		\State $\mathbf{c}^{k}_i \leftarrow \frac{\mathbf{c}^k_i}{\left \| \mathbf{c}^k_i \right \|_{2}} $
		\EndFor
		\EndFor
		
		\State initialize $\tilde{\mathbf{h}}^{k}_{u} = \mathbf{c}^k_u$
		
		\For{iteration $\tilde{t}=1$ to $\tilde{T}$}
		\For{$v \in \mathbf{N} \left( u \right)$}
		
		\State	$s^k_v(u) \leftarrow \cos(\mathbf{h}^{k}_{u}, \mathbf{c}^{k}_{v})$, $\forall k = 1, \cdots, K$.
		
		\State Calculate $p^k_{u\to v}$ and $q^k_{v \to u}$ by Eq.$\left( \ref{p} \right)$
		\EndFor
		
		\For{$k=1$ to $K$ do}
		
		\State $ {\mathbf{h}^{k}_{u}}= \tilde{\mathbf{h}}^{k}_{u} + \alpha \sum_{v \in \mathbf{N}(u)}q^{k}_{v\to u}\mathbf{c}^{k}_{v} + \beta \sum_{v \in \mathbf{N}(u)}p^{k}_{u \to v}\mathbf{c}^{k}_{v}$
		\State $\mathbf{h}^{k}_{u} =\frac{{\mathbf{h}}^{k}_{u}}{\left \| {\mathbf{h}}^{k}_{u} \right \|_{2}} $
		\State $\tilde{\mathbf{h}}^{k}_{u} \leftarrow \mathbf{h}^{k}_{u}$
		\EndFor
		\EndFor
		
		%\State $\mathbf{c}_{u} \leftarrow $ concatenation of $\{\mathbf{c}^{k}_{u}\}^{K}_{k=1}$
		\State $\mathbf{c}_{u} \leftarrow $ $\textrm{concat}(\mathbf{c}^{1}_{u},\mathbf{c}^{2}_{u},\cdots,\mathbf{c}^{K}_{u} )$
		
		%\State $\mathbf{h}_{u} \leftarrow $ concatenation of $\{\mathbf{h}^{k}_{u}\}^{K}_{k=1}$
		\State \textcolor{black}{$\mathbf{h}_{u} \leftarrow \textrm{concat}(\mathbf{h}^{1}_{u},\mathbf{h}^{2}_{u},\cdots,\mathbf{h}^{K}_{u} )$}
		
	\end{algorithmic}
	
\end{algorithm}

\subsection{Adversarial Learning for Macro-Disentanglement}
\begin{figure}[!t]	
	\centering
	\includegraphics[width=3in]{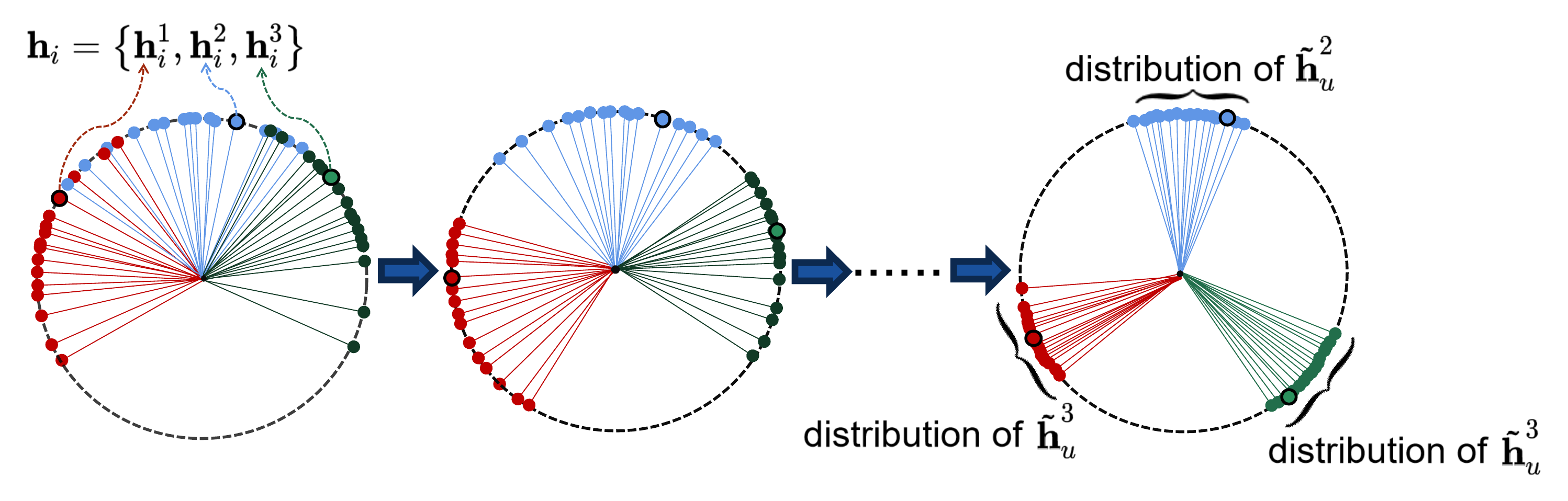}
	\caption{The iterative adversarial training tends to progressively improve the separability among component distributions.}
	\label{macro_disen}
 \vspace{-0.5cm}
\end{figure}

In essence, the goal of disentangled representation learning is to make a specific component be only sensitive to the variation in the corresponding factor, while not affected by the one in other factors~\cite{definition_dis, RDCM}.
%In other words, there should be only weak correlation or strong separability among components. Therefore, to attain the above objective,
In addition to the micro-disentanglement mentioned above, there should exist only weak correlation, or strong separability in other words, among component distributions corresponding to different factors. 

\textcolor{revise_color}{GANs have demonstrated the strong ability of distribution fitting in various fields, such as domain adaption and neural language processing~\cite{DA_survey, NLP_survey}.
Adversarial learning provides an efficient and trainable way to constrain the distances between distributions. However, adversarial learning is nontrivial to directly apply to graph disentanglement.}
Therefore, we propose a conditional adversarial regularizer for macro-disentanglement at the distribution level by explicitly imposing constraints on the dependence among components. 
%In particular, we introduce a discriminator $D$ with two kinds of output $D_{src}$ and $D_{cls}$ by weight sharing of the neural network except for the last output layers. An auxiliary classifier is added to incorporate domain information into the discriminator, which allows the single discriminator to control multiple domains (factors). Specifically, the discriminator $D$ not only discriminates whether the input sample is a disentangled component or not, but also distinguishes which component distribution it comes from.
%$K$ factors correspond to $K$ component distributions

To enable conditional adversarial learning of ADGCN, we introduce a discriminator $D$ with two branches $D_{src}$ and $D_{cls}$. Concretely, the purpose of the discrimination branch $D_{src}$ is to distinguish whether the input representation is a disentangled component representation. Relying on $D_{src}$, we have the adversarial loss term as:
\begin{equation}
\underset{\mathbf{W}}{\min}\ \underset{D}{\max}\ \mathcal{L}_{adv}=\mathbb{E}_{\mathbf{h}}[\mathrm{log}D_{src}(\mathbf{h})]+\mathbb{E}_{\mathbf{c}}[\mathrm{log}(1-D_{src}(\mathbf{c}))]
\end{equation}
Here, $\mathbf{h}^{k}_{u}$ and $\mathbf{c}^{k}_{u}$ are respectively taken as the "real" sample and the "fake" sample for disentanglement. In essence, $D$ need to maximize $\mathcal{L}_{adv}$ to distinguish between $\{\mathbf{c}^{k}_{u}\}^{K}_{k=1}$ and $\{\mathbf{h}^{k}_{u}\}^{K}_{k=1}$, while $\mathbf{W}$ will try to confuse $D$ by minimizing $\mathcal{L}_{adv}$, thus enforcing $\{\mathbf{c}^{k}_{u}\}^{K}_{k=1}$ to obey the distributions of $\{\mathbf{h}^{k}_{u}\}^{K}_{k=1}$, where $\mathbf{W}$ can be regarded as the generator $G$ in GANs. Meanwhile, due to the micro-disentanglement achieved by the dynamic assignment mechanism, $\{\mathbf{h}^{k}_{u}\}^{K}_{k=1}$ already has a certain inter-component separability. Therefore, through adversarial optimization, $\mathbf{W}$ can learn to project $\mathbf{X}$ to more distinguishable component subspaces under the guidance of minimizing $\mathcal{L}_{adv}$. In addition, it can also accelerate the convergence of Alg.~\ref{algorithm1}.

Another branch $D_{cls}$ as an auxiliary classification branch is used to identify which component subspace the input representation comes from. It is added to incorporate the factor information into the discriminator, which allows a single discriminator to control multiple factors. Relying on $D_{cls}$, we have the the component classification term as:
\begin{equation}\small
\underset{\mathbf{W}}{\min}\ \underset{D}{\min}\ \mathcal{L}_{cls}=\mathbb{E}_{k,\mathbf{h}}[-\mathrm{log}D_{cls}(k|\mathbf{h})]+\mathbb{E}_{k,\mathbf{c}}[-\mathrm{log}(D_{cls}(k|\mathbf{c}))]
\end{equation}
% It can be seen that $\mathcal{L}_{cls}$ as an explicit classification loss will guide $\mathbf{W}$ to learn component subspaces that are more easily distinguished. 
By using auxiliary classifier to minimize $\mathcal{L}_{cls}$, $D$ can learn to classify $\mathbf{c}^{k}_{\cdot}$ and $\mathbf{h}^{k}_{\cdot}$ to the corresponding $k$-th component space. $\mathbf{W}$ also tries to minimize $\mathcal{L}_{cls}$. Since $\mathbf{h}^{k}_{u}$ is derived from $\{\mathbf{c}^{k}_{i}\}_{i \in \tilde{\mathbf{N}}(u)}$, it will be beneficial of boosting the separability among component spaces and assisting $D$ to minimize $\mathcal{L}_{cls}$. 

\textcolor{revise_color}{ Due to the micro-disentanglement achieved by the dynamic assignment mechanism, $\{\mathbf{h}^{k}_{u}\}^{K}_{k=1}$ already has a certain inter-component separability. Further, with the support of the proposed conditional adversarial learning, $\mathbf{W}$ can learn to seek more distinguishable component subspaces and also accelerate the convergence of dynamic assignment through adversarial optimization. Besides, the auxiliary classification branch $D_{cls}$ provides a way to make it possible to explicitly constrain the separability between components via minimizing $\mathcal{L}_{cls}$. }
\textcolor{R2}{The discriminator with two branches for the proposed conditional adversarial learning yields a strong connection to the methods for adversarial domain adaption~\cite{UDA} with a two-branches discriminator. 
 Similarly to the one in~\cite{UDA}, $D_{src}$ in the ADGCN is also used for domain-level alignment.
 The difference is that $D_{cls}$ in~\cite{UDA} is directly related to the downstream tasks, whereas the $D_{cls}$ in the proposed conditional adversarial learning is oriented toward component classification that is not explicitly associated with the downstream task. In other words,}
\textcolor{revise_color}{the proposed conditional adversarial learning focuses on distribution alignment and component classification of representations for disentanglement instead of downstream tasks.
}
As shown in Fig.~\ref{macro_disen}, the distributions of $\{\mathbf{c}^{k}_{u}\}^{K}_{k=1}$ will become mutually independent as training progresses, i.e., $\mathbf{c}^{k}_{u}$ will obey the distribution in the specific $k$-th component space and be independent of each other.
In this way, we can capture the multiple disentangled component distributions with weak correlations to each other to achieve macro-disentanglement.

\subsection{Local Structure-aware Graph Refinement  Based on Diversity-preserving Node Sampling}

\iffalse
%\begin{figure*}[t]	
%	\centering
%	\includegraphics[width=6in]{fig/graph_refinement.png}
%	\caption{Illustration of node sampling based graph refinement. Given $\mathbf{H}^{(t)}$ that denotes the disentangled representation set of epoch $t$, we apply Gaussian kernel function to construct the global similarity matrix $S$ of $\mathbf{H}^{(t)}$. Then, We sample a subset $U$ of $\mathbf{H}^{(t)}$ from the probability calculated by Eq.(\ref{P}), and decompose $\mathbf{h}^{(t)}_{u}$ into $\{\mathbf{h}^{k}_{u}\}^{K}_{k=1}$ where $u \in U$. Further, we use $\{\mathbf{h}^{k}_{u}\}_{u \in U}$ to construct component-specific adjacency matrix $\tilde{A}^{k}$ where $k$ is from $1$ to $K$, and we merge the adjacency matrices by element-wise max pooling to obtain $\tilde{A}$. Finally, $A^{(t+1)}$ is calculated according to Eq.(\ref{AdjUpdate}).
%	}
%	\label{graph_refine}
%\end{figure*}
\fi

\begin{algorithm}[t]
	\caption{Local structure-aware graph refinement in the $t$-th epoch}
	\label{algorithm2}
        \scriptsize
	\begin{algorithmic}[1] 
		\Require $\mathcal{G} = (V, E) $, $\mathbf{H}^{(t-1)}=\left[\mathbf{h}_1^{(t-1)},\cdots \mathbf{h}_n^{(t-1)}\right]$, $\mathbf{A}^{(0)}$;
		
		\Ensure $\mathbf{A}^{(t)}$;
		\renewcommand{\algorithmicensure}{\textbf{Hyper-paramters:}}
		\Ensure: $m$, $K$, $\gamma$
		
		\State Initialization: $J_s = \phi$

		\For{$u\in V$} /* \textit{Degree-based Node Sampling} */
		
		\State $\mathbf{d}_u \leftarrow  \sum_{v}^{|V|} \mathbf{A}^{(0)}_{u,v}$
		
		\State $p_{s}(u) \leftarrow \frac{\mathbf{d}_u}{\sum_{i=1}^{n} \mathbf{d}_{i}}$
		
		\EndFor
		
		\For{$1$ to $m$}
		
		\State Take a sampling on node $u_s$ from $V$ according to $p_{s}(u_s)$
		
		\While{$v \in \tilde{\mathbf{N}}\left (u_s \right )$}
		
		\State Calculate  $\Delta_{u_{s},v} = \| \mathbf{h}_{u_s}^{(t-1)} - \mathbf{h}_{v}^{(t-1)} \|_2 $ %
		
		\State $p_{s}(v) \leftarrow p_{s}(v) \cdot f(\Delta_{u_{s},v}) $	
		
		\EndWhile
		
		\State $J_s \leftarrow J_s \cup \{u_s\}$
		
		\EndFor

		\While{sample node $u_s \in J_s$} /* \textit{Graph Refinement} */
		
		\For{$k=1$ to $K$}
		
		\State Select $\mathcal{k}$ nearest neighbors of $\mathbf{h}^{(t-1),k}_{u_s}$ as ${\mathbf{N}'}\left (u_s \right )$
		
		\State $ \tilde{\mathbf{A}}^{(t-1),k}_{u_s,{\mathbf{N}'}\left (u_s \right )} \leftarrow \cos (\mathbf{h}^{(t-1),k}_{u_s}, \mathbf{h}^{(t-1),k}_{{\mathbf{N}'}\left (u_s \right )})$
		%, $\forall v \in {\mathbf{N}'}\left (u_s \right )$
		
		\EndFor
		
		\EndWhile
		
		\State $\tilde{\mathbf{A}}^{(t-1)} \leftarrow \textrm{maxpooling}(\tilde{\mathbf{A}}^{(t-1),1},\tilde{\mathbf{A}}^{(t-1),2}, \cdots, \tilde{\mathbf{A}}^{(t-1),K})$

		\State Obtain the refined $\mathbf{A}^{(t)}$  by Eq.(\ref{refining}).
	\end{algorithmic}
\label{refinement}
\end{algorithm}

\begin{figure*}[t]	
	\centering
	\includegraphics[width=6in]{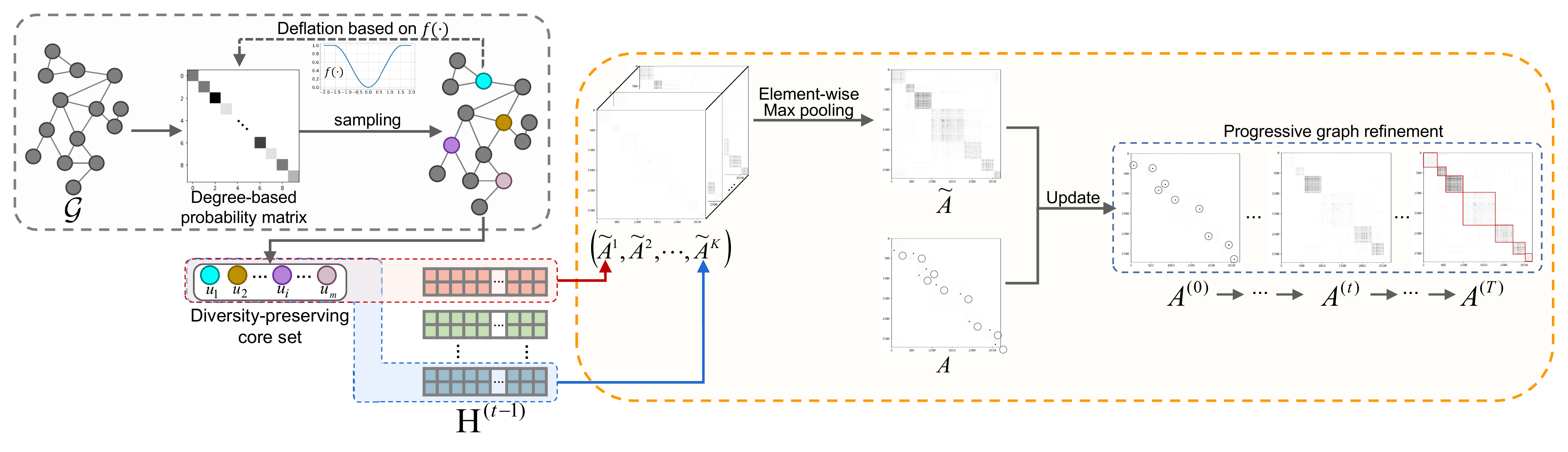}
	\vspace{-0.3cm}
	\caption{Illustration of the diversity-preserving graph refinement. Since the initially given adjacency matrix $\mathbf{A}$ is extremely sparse, we have marked the existing edges $\mathbf{A}_{ij}$ using a circle for clearer presentation.}
	\vspace{-0.5cm}
	\label{graph_refine}
\end{figure*}
For real-world graph data, the complex relationship between nodes is often represented as a hard binary link (i.e., 0 or 1). Obviously, it is a discrete and simplified form of continuous relationship between nodes, which seriously limits the expressibility of the learned node representation. On the other hand, the node representation obtained in the embedding space can in turn be used to reveal the intrinsic relationship between nodes. To better characterize the node relationships and further facilitate the learning of node representation, an intuitive way is to refine the originally given graph structure with the embedded node representations~\cite{IDGL}. However, such global refinement of the relationships among all nodes without distinction will inevitably lead to some noisy edges, which may further confuse the training of the node representation learning model. In addition, it also has scalability problems on large graphs. To address these issues,  we propose a local structure-aware graph refinement as shown in Fig.~\ref{graph_refine} to progressively reveal the latent relationships of nodes, thus achieving efficient and robust graph refinement.

The key to achieving local structure-aware graph refinement is how to obtain a diversified core subset $J_s \subset V$ for local refinement, so as to maintain the global coverage of the graph as much as possible. Determinant Point Process (DPP)~\cite{DPP} is a classical algorithm for diversity-preserving sample selection. 
However, the high computational cost of DPP is somewhat the opposite of what we intended. Hence, for local structure-aware graph refinement, we propose a degree-based node sampling method with good diversity preservation to serve as an efficient approximation of DPP  sampling on graph data.

Considering the characteristics of the graph structure, the normalized degree centrality indicating the importance of a node is used as the sampling probability $p_s(\cdot)$ for node selection, that is to say, we have $p_s(u)=\frac{\mathbf{d}_{u}}{\sum_{i=1}^{n}\mathbf{d}_{i}}$ for $u$-th node. Let $J_{s}\subset{V}$, $|J_s| = m$, denote the core set to be selected for local aware graph refinement.   
To realize the diversity of the selected core set $J_s$, we propose a sequential node selection method by means of soft-deflation. Specifically, for a node $u_s\in J_s$ that has been selected, a soft-deflation is applied to its every neighbor $v \in \tilde{\mathbf{N}}\left (u_s \right )$, i.e., we will have $p_s(v) \cdot f(\Delta_{u_{s},v} )\rightarrow{p_s(v)}$, where $f(\cdot)\in [0,1)$ is the soft-deflation function and $\Delta_{u_{s},v}$ denotes the distance between $u_s$ and its neighbor $v$.
Here, we follow the definition of $f(\cdot)$ in~\cite{DPP-landmark} and consider $f(\Delta) = \sin^2(\Delta/\tau)$, where $\tau$ is the scaling factor to keep $\Delta/\tau$ ranging in $[0, \pi/2)$. 
Through the soft-deflation in sequential node selection, we can ensure that, for each node in the associated neighborhood node set $\Psi(J_s) = \bigcup_{u_s \in J_s}\tilde{\mathbf{N}}(u_s)$, it will be further selected into the core set with less probability in the subsequent node selection. In this way, it means that the finally obtained core set $J_s$ can achieve the maximum coverage of the graph to a certain extent.

Then, with the assistance of the diversity-preserving core subset $J_s$, the local structure-aware graph refinement can be carried out as follows:
\begin{align}
%A^{(t)}_{u_s,v\in \tilde{\mathbf{N}}\left (u_s \right )}=  (1 - \gamma) A^{0}_{u_s,v \in \tilde{\mathbf{N}}\left (u_s \right )} + \gamma \tilde{A}_{u_s, v \in \tilde{\mathbf{N}}\left (u_s \right )}^{(t)}
\mathbf{A}^{(t)}_{u_s,{\mathbf{N}}\left (u_s \right )}=  (1 - \gamma) \mathbf{A}^{(0)}_{u_s,{\mathbf{N}}\left (u_s \right )} + \gamma \tilde{\mathbf{A}}^{(t-1)}_{u_s, {\mathbf{N}}\left (u_s \right )}
\label{refining}
\end{align}
where $u_s \in J_s$, $\mathbf{A}^{(0)}$ represents the initial adjacency matrix and $\tilde{\mathbf{A}}^{(t-1)}$ denotes the adjacency refining matrix in the $t$-th epoch, and $\gamma$ is a trade-off coefficient. 
%Obviously, Eq.\ref{refining} is a local refinement based on $u_s$, but the neighbors of all the nodes in $J_s$ can be define as a neighbor set:
%\begin{align}
%\Psi(J_s) = \bigcup_{u_s \in J_s}{\mathbf{N}}(u_s) 
%\end{align}
%where the neighbor set $\Psi(J_s)$ is the the maximum subset of coverage of $V$ based on $J_s$. 
%

 The link relationship between two nodes may come from a certain component space. 
To capture the underlying topological structure of the graph, we first obtain $K$ component adjacency matrices $\{\tilde{\mathbf{A}}^{(t-1),k}\}_{k=1}^{K}$ based on the associated node representations $\{\mathbf{h}^{(t-1),k}_{u}\}_{u \in V}$ in $t$-th epoch, and then an element-wise max pooling is applied to form a component sensitive adjacency matrix $\tilde{\mathbf{A}}^{(t-1)} = \mathbf{max pooling}(\tilde{\mathbf{A}}^{(t-1),1},\tilde{\mathbf{A}}^{(t-1),2}, \cdots, \tilde{\mathbf{A}}^{(t-1),K})$. The adopted max pooling plays the role of integrating the specific structures of multiple latent components into a unified topological one, boosting the link characteristics of nodes.

Actually, it can be found from Eq.(\ref{refining}) that the local aware refinement guided by a core set $J_s$ can spread to the overall graph, thus achieving global refinement of the graph structure. Meanwhile, it can not only prevent the introduction of noisy edges in the refinement of the graph but also help to learn more robust disentangled representations progressively with the progress of training.  \textcolor{black}{Due to the frequent graph refinement, it may lead to an unstable training process. Therefore, in our practice, we perform graph refinement only in epochs when the model achieves better performance than before in the validation set.} The detail about local aware graph refinement with diversity-preserving node sampling is illustrated in \textbf{Alg.}~\ref{refinement}.

\section{Optimization}

As we mainly focus on the task of node classification in this paper, the last layer of ADGCN is designed as a full-connected layer to output the predicted class label of nodes. Besides, ReLU is used as activation to obtain the component representations $\{{\mathbf{c}^{k}_u}\}_{k=1}^{K}$ after projections. 
Specifically, the overall loss of ADGCN is:
\begin{equation}
\mathcal{L} = \mathcal{L}_{t} +  \lambda (\mathcal{L}_{adv} + \eta \mathcal{L}_{cls})\ , \ \mathcal{L}_{D} =  -\mathcal{L}_{adv} + \eta \mathcal{L}_{cls} 
\label{loss_G}
\end{equation}
% \begin{equation}
% \mathcal{L}_{D} =  -\mathcal{L}_{adv} + \eta \mathcal{L}_{cls}
% \label{loss_D}
% \end{equation}
where $\mathcal{L}_{t}$ denotes the task-aware loss, $\lambda$ and $\eta$ are hyper-parameters of the adversarial regularizer. For the semi-supervised node classification task, $\mathcal{L}_{t}$ is the cross-entropy loss. The same as~\cite{DisenGCN}, we use the multi-label soft margin loss as $\mathcal{L}_{t}$ for the multi-label node classification task:
\begin{align}
\mathcal{L}_{t} = -\frac{1}{C}\sum ^{C}_{c=1} &[\mathbf{y}_{u}(c) \log [g(\hat{\mathbf{y}}_{u}(c))] \nonumber
\\ &+(1-\mathbf{y}_{u}(c)) \log[g(-\hat{\mathbf{y}}_{u}(c))]]
\end{align}
where $g(\cdot)$ denotes the $\mathbf{sigmoid}$ function, $\mathbf{y}_{u}(c)$ and $\hat{\mathbf{y}}_{u}(c)$ denote the true label and the predicted label of node $u$, respectively. We optimize the proposed model with Adam optimizer~\cite{Adam}. 

Since $\mathbf{h}^{k}_{u}$ and $\{\mathbf{c}^{k}_{i}\}_{i \in \tilde{\mathbf{N}}(u)}$ are dynamically correlated, \textcolor{revise_color}{the conventional GAN training method~\cite{GAN} is not suitable for such case since there are no generated samples and real disentangled samples for graph disentanglement as the “fake” and “real” samples in conventional GANs or the "source domain" and "target domain" in the field of domain adaption.} Thus, an iterative adversarial training is applied to progressively improve the separability among component spaces via an `EM-like' procedure given by \textbf{Alg}.~\ref{EM-like}. \textcolor{black}{Here, the expression of `EM-like' refers to the iterative training process that is similar to EM. In fact, the process of aggregating $\{\mathbf{c}_{i}\}^{n}_{i=1}$ into $\{\mathbf{h}_{i}\}^{n}_{i=1}$ is equivalent to the E-step, where $\{\mathbf{h}_{i}\}^{n}_{i=1}$ can be seen as the achieved  `expectation'. Then, the separability between components is expected to be maximized via adversarial learning, i.e., the `maximization' of the M-step.}

\begin{algorithm}[t]
	\caption{EM-like procedure of the proposed ADGCN}
	\label{algorithm3}
	\footnotesize
	\begin{algorithmic}[1] %Ã¿ÐÐÏÔÊ¾ÐÐºÅ
		\Require
		$\mathcal{G} = \{V, E\} $, $\mathbf{X}$, $A$.
		
		%$\left \{ \mathbf{x}_i \right \} \in \mathbb{R}^f$: the feature vector of node, $i \in V$
		
		%$K$: the number of component
		
		%$T$: the number of iterations of training
		
		%$\tilde{T}$: the number of iterations of dynamic assignment
		
		\Ensure $\mathbf{H}$: the node disentangled representation matrix of the given graph
		\renewcommand{\algorithmicensure}{\textbf{Hyper-paramters:}}
		\Ensure:  $K$, $\tilde{T}$, $T$, $m$, $\gamma$, $\lambda$, $\eta$
		
		\State Initialization:
		
		$\mathbf{W}=\{\mathbf{W}_k\}_{k=1}^{K}$: projection matrix.
		
		\textcolor{black}{$\alpha$, $\beta$: coefficients of component-specific aggregation.}
		
		$\theta_{D}$: parameters of the discriminator $D$.

		%\Ensure $\mathbf{h}_u$: the disentangled representation of node $u$
		\For{epoch $t = 1$ to $T$}
		
		\State \textbf{E-step:}
		
		Obtain $\left \{ \mathbf{h}_i^{(t-1)} \right \}_{i=1}^n$ and $\left \{ \mathbf{c}_i ^{(t-1)}\right \}_{i=1}^n$ through \textbf{Alg.}~\ref{algorithm1}.
		
		Obtain $A^{(t)}$ through \textbf{Alg.}~\ref{algorithm2} if performing graph refinement.
		
		\State \textbf{M-step:}
		
		Sample the "real" components from $\left \{ \mathbf{h}_i^{(t-1)} \right \}_{i=1}^n$.
		
		Sample the "fake" components from $\left \{ \mathbf{c}_i^{(t-1)} \right \}_{i=1}^n$.
		
		Update $\mathbf{W}_k$, $k=1,\cdots,K$ by minimizing $\mathcal{L}$ in Eq. (\ref{loss_G}).
		
		Update $\theta_{D}$ by minimizing $\mathcal{L}_{D}$ in Eq. (\ref{loss_G}).
		
		\EndFor
		
	\State \textbf{return} $\mathbf{H}=\left[\mathbf{h}_1^{(T-1)},\cdots,\mathbf{h}_n^{(T-1)}\right]$.
	\end{algorithmic}
\label{EM-like}	
\end{algorithm}

\begin{table}[t]\scriptsize
\vspace{-0.2cm}
	\centering
	\caption{Statistics of the used datasets.}
	\begin{tabular}{lrrrrc}
		\hline
		Dataset                           & \multicolumn{1}{c}{Nodes} & \multicolumn{1}{c}{Edges} & \multicolumn{1}{c}{Classes} & \multicolumn{1}{c}{Features}  & Multi-label \\ \hline
		Citeseer~\cite{Citeseer}               & 3327                      & 4552                      & 6                           & 3703                                             & No          \\
		Cora~\cite{Cora}                   & 2708                      & 5278                      & 7                           & 1433                                             & No          \\
		Pubmed~\cite{Pubmed}                 & 19717                     & 44324                     & 3                           & 500                                              & No          \\
		CS~\cite{pitfall}       & 18333                     & 81894                     & 15                          & 6805                                             & No          \\
		Computers~\cite{pitfall}    & 13381                     & 245778                    & 10                          & 767                                              & No          \\
		Photo~\cite{pitfall}        & 7487                      & 119043                    & 8                           & 745                                              & No          \\ \hline
		Wikipedia~\cite{POS_tagger}                  & 4777                      & 184812                    & 40                          & -                                                & Yes         \\
		Blogcatalog~\cite{blog}              & 10312                     & 333983                    & 39                          & -                                                & Yes         \\ \hline
	\end{tabular}
	\label{table::dataset}
	\vspace{-0.2cm}
\end{table}

\begin{table*}[t]\scriptsize
	\centering
	\caption{ Results on citation datasets with both fixed and random splits for semi-supervised node classification accuracy ($\%$).Boldface letters mark the best result, while underlined letters denote the second-best result.}
	\vspace{-0.2cm}
\begin{tabular}{l|l|cccccc}
\hline
 &
  \multirow{2}{*}{Models} &
  \multicolumn{2}{c}{Cora} &
  \multicolumn{2}{c}{CiteSeer} &
  \multicolumn{2}{c}{PubMed} \\ \cline{3-8} 
 &
   &
  Fixed &
  Random &
  Fixed &
  Random &
  Fixed &
  Random \\ \hline
  &
  MLP &
  $61.6\pm0.6$ &
  $59.8\pm2.4$ &
  $61.0\pm1.0$ &
  $58.8\pm2.2$ &
  $74.2\pm0.7$ &
  $70.1\pm2.4$ \\ \hline
\multirow{3}{*}{Spectral-based} &
  ChebNet~\cite{ChebNet} &
  $80.5\pm1.1$ &
  $76.8\pm2.5$ &
  $69.6\pm1.4$ &
  $67.5\pm2.0$ &
  $78.1\pm0.6$ &
  $75.3\pm2.5$ \\
 &
  GCN~\cite{GCN} &
  $81.3\pm0.8$ &
  $79.1\pm1.8$ &
  $71.1\pm0.7$ &
  $68.2\pm1.6$ &
  $78.8\pm0.6$ &
  $77.1\pm2.7$ \\
 &
 ARMA~\cite{ARMA} &
  $83.4\pm0.6$ &
  $79.7\pm1.9$ &
  $72.5\pm0.4$ &
  $68.9\pm2.2$ &
  $78.9\pm0.3$ &
  $78.4\pm2.0$ \\ \hline
\multirow{4}{*}{Spatial-based} &
  APPNP~\cite{PPNP} &
  $83.3\pm0.5$ &
  $\textbf{81.9}\pm\textbf{1.4}$ &
  $71.8\pm0.4$ &
  $69.8\pm1.7$ &
  $80.1\pm0.2$ &
  \underline{$79.5\pm2.2$} \\
 &
  SGC~\cite{SGC} &
  $81.7\pm0.1$ &
  $80.4\pm1.8$ &
  $71.3\pm0.2$ &
  $68.7\pm2.1$ &
  $78.9\pm0.1$ &
  $76.8\pm2.6$ \\
 &
  GAT~\cite{GAT} &
  $83.1\pm0.4$ &
  $80.8\pm1.6$ &
  $70.8\pm0.5$ &
  $68.9\pm1.7$ &
  $79.1\pm0.4$ &
  $77.8\pm2.1$ \\
  &
  AP-GCN~\cite{AP-GCN} &
  $82.7\pm0.3$ &
  $79.8\pm3.2$ &
  $73.3\pm0.6$ &
  $69.2\pm3.0$ &
  $80.1\pm0.7$ &
  $79.2\pm1.6$ \\ 
  \hline
\multirow{6}{*}{Disen-based} &
DisenGCN~\cite{DisenGCN} &
  $83.1\pm1.1$ &
  $80.3\pm2.0$ &
  $73.6\pm0.8$ &
  $70.1\pm1.8$ &
  $79.7\pm0.7$ &
  $78.7\pm1.7$ \\
 &
  IPGDN~\cite{IPGDN} &
  $84.1$ &
  - &
  $74.0$ &
  - &
  $\textbf{81.2}$ &
  - \\ 
 &
  FactorGCN~\cite{FactorGCN} &
  $82.6\pm1.0$ &
  $78.1\pm1.8$ &
  $72.2\pm0.9$ &
  $69.1\pm1.0$ &
  $79.4\pm1.1$ &
  $77.8\pm2.1$ \\
 &
  SGCN~\cite{SGCN} &
  $82.2\pm0.8$ &
  $79.9\pm1.5$ &
  $73.6\pm0.9$ &
  $69.5\pm1.1$ &
  $80.3\pm1.5$ &
  $79.3\pm0.7$ \\
 &
  LGD-GCN~\cite{LGD-GCN} &
  $84.2\pm0.4$ &
  $79.6\pm2.1$ &
  \underline{$74.3\pm0.3$} &
  $70.3\pm1.6$ &
  $80.3\pm1.5$ &
  \underline{$80.9\pm0.4$} \\
 &
  DisGNN~\cite{DisGNN} &
  $82.8\pm0.9$ &
  $80.0\pm1.9$ &
  $73.4\pm0.3$ &
  $68.5\pm1.7$ &
  $80.3\pm0.2$ &
  $79.1\pm1.1$ \\ \cline{2-8} 
 &
  \textbf{ADGCN}(Ours) &
  \underline{$84.3\pm0.6$} &
  $80.8\pm1.7$ &
  \underline{$74.3\pm0.7$} &
  \underline{$70.4\pm1.5$} &
  $\textbf{81.2}\pm\textbf{0.5}$ &
  $79.2\pm1.7$ \\
 &
  \textbf{ADGCN-R}(Ours) &
  $\textbf{84.5}\pm\textbf{0.7}$ &
  \underline{$81.0\pm2.0$} &
  $\textbf{74.5}\pm\textbf{0.8}$ &
  $\textbf{71.0}\pm\textbf{1.4}$ &
  ${80.4}\pm{0.4}$ &
  $\textbf{79.9}\pm\textbf{1.3}$ \\ \hline
 & 
  \textbf{P-value} 
  & $3.43 \times 10^{-2}$
  & $2.21 \times 10^{-2}$
  & $6.82 \times 10^{-4}$
  & $1.17 \times 10^{-18}$
  & $4.45 \times 10^{-21}$
  & $1.48 \times 10^{-5}$ \\ \hline
\end{tabular}
\label{citation}
\vspace{-0.3cm}
\end{table*}

\section{Experimental Results and Analysis}
%In this section, we detail our experimental protocol and then present comparison results of ADGCN with the state-of-the-art methods on node classification tasks.
\subsection{Evaluation Setup}

\begin{table}[t]\scriptsize
	\caption{ Results on the other three datasets with random splits in terms of semi-supervised node classification accuracy ($\%$).}
	\vspace{-0.2cm}
	\begin{tabular}{lccc}
		\hline
		Models &
		\begin{tabular}[c]{@{}c@{}}CS\end{tabular} &
		\begin{tabular}[c]{@{}c@{}}Computers\end{tabular} &
		\begin{tabular}[c]{@{}c@{}}Photo\end{tabular} \\ \hline
		MLP                                 & $88.3\pm0.7$                  & $44.9\pm5.8$                  & $69.6\pm3.8$                  \\
		LabelProp~\cite{lp}                 & $73.6\pm3.9$                  & $70.8\pm8.1$                  & $72.6\pm11.1$                 \\
		LabelProp NL~\cite{lp}              & $76.7\pm1.4$                  & $75.0\pm2.9$                  & $83.9\pm2.7$                  \\
		GCN~\cite{GCN}                      & $91.1\pm0.5$                  & $82.6\pm2.4$                  & $91.2\pm1.2$                  \\
		ARMA~\cite{ARMA}                    & $92.4\pm0.3$                  & $80.7\pm0.3$                  & $89.4\pm0.3$                  \\
		GAT~\cite{GAT}                      & $90.5\pm0.6$                  & $78.0\pm19.0$                 & $85.7\pm20.3$                 \\
		MoNet~\cite{MoNet}                  & $90.8\pm0.6$                  & $83.5\pm2.2$                  & $91.2\pm1.3$                  \\
		SAGE-mean~\cite{Graphsage}     & $91.3\pm2.8$                  & $82.4\pm1.8$                  & $91.4\pm1.3$                  \\
		SAGE-maxpool~\cite{Graphsage}  & $85.0\pm1.1$                  & -                             & $90.4\pm1.3$                  \\
		SAGE-meanpool~\cite{Graphsage} & $89.6\pm0.9$                  & $79.9\pm2.3$                  & $90.7\pm1.6$                  \\
		APPNP~\cite{PPNP}                   & $93.2\pm0.1$                  & $80.1\pm0.3$                  & $89.3\pm0.2$                  \\
		AP-GCN~\cite{AP-GCN}                & $92.2\pm0.2$                  & $82.1\pm0.3$                  & $91.3\pm0.9$                  \\
		% DisenGCN~\cite{DisenGCN}            & $92.9\pm 0.2$                 & $84.6\pm0.9$                  & $91.7\pm0.7$                  \\
		FactorGCN~\cite{FactorGCN}          & $90.9\pm2.3$                  & $82.1\pm1.5$                  & $88.2\pm3.1$                  \\
            SGCN~\cite{SGCN}                    & $91.1\pm1.7$                  & $84.1\pm1.1$                  & $90.6\pm1.2$                  \\
            LGD-GCN~\cite{LGD-GCN}              & $92.7\pm0.3$                  & $83.8\pm1.4$                  & \underline{$91.8\pm0.3$}      \\
		DisGNN~\cite{DisGNN}                & OOM                           & OOM                           & OOM                           \\  \hline
		\textbf{ADGCN}(Ours)                & \underline{$92.9 \pm 0.3$}    & \underline{$84.7\pm1.1$}      & $91.7\pm0.4$ \\
		\textbf{ADGCN-R}(Ours)              & $\textbf{93.3}\pm0.4$         & $\textbf{85.2}\pm0.7$         & $\textbf{92.1}\pm0.3$         \\  \hline
            \textbf{P-value}                    & $1.74 \times 10^{-5}$         & $2.10 \times 10^{-5}$         & $2.71 \times 10^{-10}$         \\ \hline
	\end{tabular}
\label{co-network}
\vspace{-0.4cm}
\end{table}
\subsubsection{Datasets}
We adopt eight real-world graph datasets and the statistics about them are summarized in Table \ref{table::dataset}.
On the evaluation of the semi-supervised node classification and clustering, three citation networks Cora~\cite{Cora}, Citeseer~\cite{Citeseer}, and Pubmed~\cite{Pubmed} are used, whose links and labels represent citations and research fields, respectively. In addition,  we also take three co-authorship and co-purchase graphs, i.e., Coauthor CS~\cite{pitfall}, Amazon Computers~\cite{pitfall}, and Amazon Photo~\cite{pitfall}.
%Cora~\cite{Cora}, Citeseer~\cite{Citeseer}, and Pubmed~\cite{Pubmed} are citation networks for the semi-supervised node classification task, whose links represent citations and labels denote different research fields.
For multi-label node classification, Wikipedia~\cite{node2vec} and Blogdatalog~\cite{blog} are used.  
Wikipedia is a co-occurrence network whose labels represent the Part-of-Speech (POS) tags.
Blogdatalog is a social network, whose labels and links represent user interests and social relations respectively. Since these two datasets do not contain node features, we use the rows of the adjacency matrix as node features.

\begin{table*}\scriptsize
  \begin{minipage}[p]{0.35\textwidth} 
  \caption{\textcolor{revise_color}{Evaluation of Adversarial Learning (\%).}}
	\setlength\tabcolsep{3pt}
	\centering
	\scriptsize
	\begin{tabular}{ccccc}
		\toprule
		\textbf{Datasets} & \textbf{Objective}  & \textbf{w/o $\mathcal{L}_{cls}$}  & \textbf{w/o $\mathcal{L}_{adv}$}  & \textbf{ADGCN}\\
		\midrule
		\multirow{3}*{Cora}
		& WGAN-GP      & 80.8  & 80.2 & 81.4\\
		& LSGAN        & 79.4  & 80.2 & 80.3 \\
		& Vanilla GAN  & 80.3  & 80.2 & 80.8 \\ \hline
		\multirow{3}*{Citeseer}
		& WGAN-GP      & 70.2  & 69.1 & 70.8\\
		& LSGAN        & 70.0  & 69.1 & 70.4\\
		& Vanilla GAN  & 69.6  & 69.1 & 70.4 \\ \hline
		\multirow{3}*{Pubmed}
		& WGAN-GP      & 78.4  & 78.3 & 78.9\\
		& LSGAN        & 79.1  & 78.3 & 79.4\\
		& Vanilla GAN  & 78.9  & 78.3 & 79.2\\
		
		\bottomrule
	\end{tabular}
        \label{exp::adversarial_learning}
  \end{minipage}
\begin{minipage}[p]{0.7\textwidth}
\caption{Node clustering results(\%).}
	\setlength\tabcolsep{2pt}
	\centering
	\scriptsize
	\begin{tabular}{ccccccccc}
		\toprule
		\textbf{Datasets} & \textbf{Metrics}  & \textbf{ChebNet}\cite{ChebNet} & \textbf{GCN}\cite{GCN}  & \textbf{GAT}\cite{GAT} & \textbf{DisenGCN}\cite{DisenGCN}& \textbf{IPGDN}\cite{IPGDN} & \textbf{FactorGCN}\cite{FactorGCN} & \textbf{ADGCN}\\
		\midrule
		\multirow{3}*{Cora}
		& ACC  & 71.9 & 73.5  & 75.2  & 75.5  & 76.1 & 75.1    & \textbf{81.0}$\pm$1.5 \\
		& NMI  & 49.8 & 51.7  & 57.0  & 58.4  & 59.2 & 58.0    & \textbf{62.3}$\pm$1.0 \\
		& ARI  &42.4  &48.9   &54.1   & 60.4  & 61.0 & 59.3    & \textbf{62.1}$\pm$1.2\\ \hline
		\multirow{3}*{Citeseer}
		& ACC  & 65.0 & 67.7  & 68.0  & 68.2  & 68.9 & 66.9    & \textbf{69.3}$\pm$0.5\\
		& NMI  & 42.6 & 42.8  & 43.1  & 43.7  & 44.3 & 40.7    & \textbf{46.6}$\pm$0.5\\
		& ARI  &41.5  &42.8   &43.6   & 42.5  & 43.0 & 42.2    & \textbf{47.6}$\pm$1.1\\ \hline
		\multirow{3}*{Pubmed}
		& ACC  & 75.2 & 75.6  & 76.3  & 77.0  & 77.8 & 77.2   & \textbf{78.4}$\pm$0.7\\
		& NMI  & 35.6 & 35.0  & 35.0  & 36.1  & 37.0 & 36.9   & \textbf{39.2}$\pm$1.3\\
		& ARI  &38.6  &40.9   &41.4   & 41.6  & 42.0 & 41.4   & \textbf{44.1}$\pm$0.9\\
		
		\bottomrule
	\end{tabular}
	\label{node_clustering}
\end{minipage}
 \vspace{-0.6cm}
\end{table*}

\begin{figure*}[t]
	
	\subfigure[Macro-F1($\%$), Wikipedia]{
		\centering
		\includegraphics[width=1.5in]{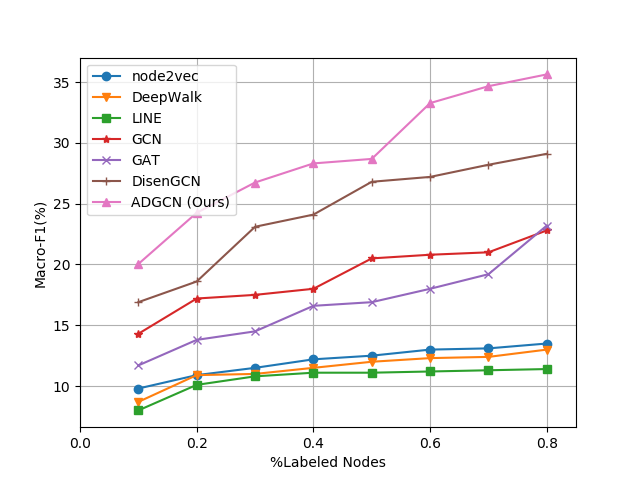}
		%\caption{fig1}
	}%
	\subfigure[Micro-F1($\%$), Wikipedia]{
		\centering
		\includegraphics[width=1.5in]{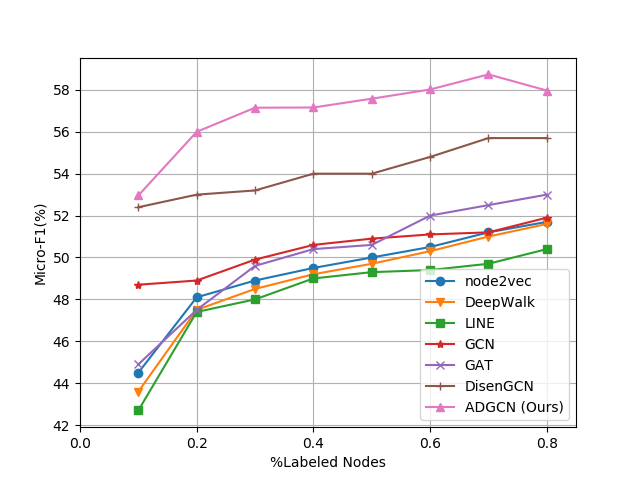}
		%\caption{fig1}
	}%
\centering
	\subfigure[Macro-F1($\%$), Blogdatalog]{
		\centering
		\includegraphics[width=1.5in]{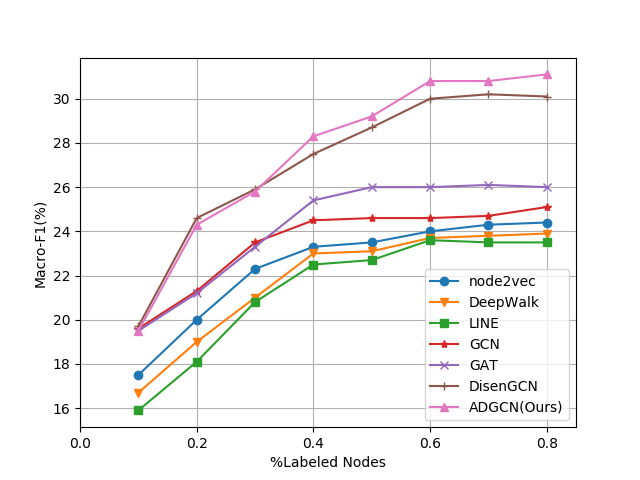}
		%\caption{fig2}
	}%
	%\\
	\subfigure[Micro-F1($\%$), Blogdatalog]{
		\centering
		\includegraphics[width=1.5in]{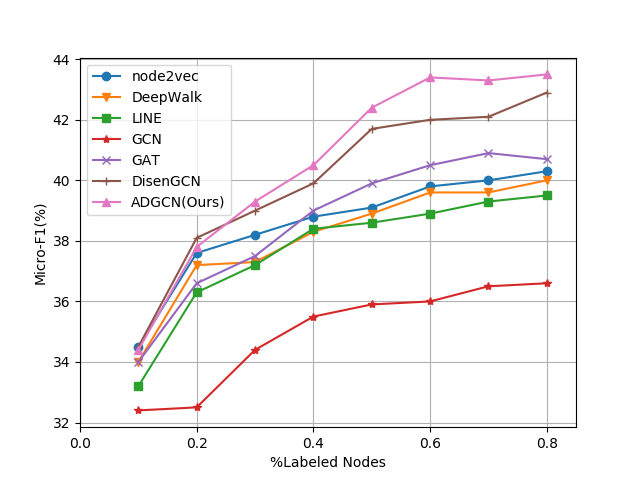}
		%\caption{fig2}
	}%
	\centering
	\caption{Macro-F1 and Micro-F1 scores on the multi-label classification tasks.}
	\centering
	\label{multi-classificaion}
	\vspace{-0.5cm}
\end{figure*}

\subsubsection{Comparison methods}

We select several baselines for comparison: (i) general baselines, including MLP, Label Propagation (LabelProp)~\cite{lp}, Planetoid~\cite{planetoid}; (ii) spectral convolution-based methods, including ChebNet~\cite{ChebNet}, GCN~\cite{GCN}, and ARMA~\cite{ARMA}; (iii) spatial convolution-based methods, including GAT~\cite{GAT}, MoNet~\cite{MoNet}, GraphSAGE~\cite{Graphsage}, APPNP~\cite{PPNP}, SGC~\cite{SGC}, and AP-GCN~\cite{AP-GCN}. Besides, we also compare the most related state-of-the-art works that are focused on graph disentanglement in graph representation learning, \textcolor{black}{i.e., DisenGCN~\cite{DisenGCN}, IPGDN~\cite{IPGDN}, FactorGCN~\cite{FactorGCN}, SGCN~\cite{SGCN}, LGD-GCN~\cite{LGD-GCN}, and DisGNN~\cite{DisGNN}.} Due to the good performance on multi-label node classification, three probabilistic models DeepWalk~\cite{deepwalk}, LINE~\cite{LINE}, and node2vec~\cite{node2vec} are also used for comparison.
%It is worth noting that the open source implementations of GCN and GAT cannot perform multi-label classification, For fair comparison, we follow the modification made to GCN and GAT in~\cite{DisenGCN}.

\subsubsection{Implementation details}

To detail the implementation, we define $\Delta d = \frac{d}{K}$ as the dimension of node components, and $K^{(l)}$ as the number of components in the $l$-th layer, respectively. And $\tilde{T}$ denotes the iterations of the dynamic assignment in the component-specific aggregation layer.

In semi-supervised node classification,
we fix $K^{(l)}$ to $K=5$ for the adopted six datasets and the dimension of $\mathbf{h}_{u}$ is set to be $K \cdot \Delta d$. For the multi-label node classification task, due to a large number of tags, we follow the setting in DisenGCN. Specifically, the number $K$ of the intermediate layers is set to gradually decrease to learn the hierarchical disentangled representation, and we use skip-connection to preserve the representation of different levels. As for the output dimension of the first layer, it is set to $K^{(1)} \cdot \Delta d = 128$, where $K^{(1)} \in \{4,8,\cdots,32\}$, $K^{(l)}-K^{(l+1)}=\Delta K \in \{0,2,4\}$, and $\Delta d$ is fixed to $\frac{128}{K^{(1)}}$. As hyper-parameters, both $K^{(1)}$ and $\Delta K$ are searched through hyperopt~\cite{hyperopt}. 
For other hyper-parameters, both $\lambda$ and $\eta$ are set to 1 empirically and we tune the remaining hyper-parameters of ADGCN using hyperopt~\cite{hyperopt}. For each hyper-parameter combination, we run 200 epochs and choose the best combination on the validation set. Then we report the averaged results of 30 times of execution on each dataset. 
%, including the number of layers, the learning rate, the dropout rate, and the $l_2$ regularization term

%\footnotemark[1] for IDGL
%\footnotetext[1]{On the phase of manuscript submission, we have noticed a very recent SOTA model IDGL and our model achieved .}
\vspace{-0.2cm}
\subsection{Node Classification}
\subsubsection{Semi-Supervised Node Classification}
In view of semi-supervised node classification on Cora, Citeseer, and Pubmed datasets, we use the same fixed dataset split as GCN with each dataset containing only 20 labeled nodes for each class, 500 nodes for validation, and 1000 nodes for testing. Meanwhile, we also conduct a random split as~\cite{PPNP}, and the random split has the same ratio as the fixed split. For Coauthor CS, Amazon Computers, and Amazon Photo, the random split is also adopted, that is, 20 labeled nodes per class are taken as the training set, 30 labeled nodes per class nodes as the validation set, and the rest nodes as the test set.

As shown in Table \ref{citation} and \ref{co-network}, the results with the best average performance are bolded and ADGCN-R denotes ADGCN with graph refinement.
%This proves the effectiveness of graph disentanglement to a certain extent.
\textcolor{black}{Due to the fact that FactorGCN is primarily devoted to graph-level disentanglement instead of node-level disentanglement, the performance of FactorGCN for node classification is unsatisfactory.}  Compared with the holistic approaches, such as GAT and GCN, DisenGCN, IPGDN, and ADGCN achieve better performance. In particular, ADGCN outperforms over DisenGCN by about $0.8\%$, $1.0\%$, and $0.9\%$ on Cora, Citeseer, and Pubmed datasets, respectively. Besides, ADGCN-R also achieves the best performance on three co-author/purchase datasets. \textcolor{black}{Meanwhile, the standard deviation of the results for ADGCN and ADGCN-R are smaller than that for DisenGCN, indicating that the performance of the proposed method is more stable as compared to DisenGCN.}
%This just verifies that ADGCN can further boost the performance, which benefits from the combination of macro-disentanglement and micro-disentanglement.
\textcolor{black}{Furthermore, to verify the improvements achieved by ADGCN-R, we also perform a paired t-test between ADGCN and ADGCN-R with respect to the accuracy.   The reported p-values in Tables~\ref{citation} and \ref{co-network} show that the results on different test datasets are statistically significant with p $<$ 0.05 via the paired t-test.}

\subsubsection{Multi-label Node Classification}
Following the experiment settings in~\cite{node2vec}, we report the performance of each method by increasing the number of nodes labeled for training from 10 $\%$ to 80 $\%$ of $\left | V \right | $, and the other nodes are divided into two sets for verification and testing.
Macro-F1 and Micro-F1 are used to evaluate the performance of each model. As  can be observed  from  Fig.~\ref{multi-classificaion}, ADGCN performs the best compared to all baselines including DisenGCN, especially on the Wikipedia dataset. Meanwhile, in terms of Macro-F1, graph disentanglement methods achieve a higher score than the others. It indicates that graph disentanglement can facilitate the understanding of the relation between nodes, so as to be more robust to graph data with class imbalance.

\begin{figure}[t]
	\centering
	\subfigure[Cora]{
		\centering
		\includegraphics[width=1.3in]{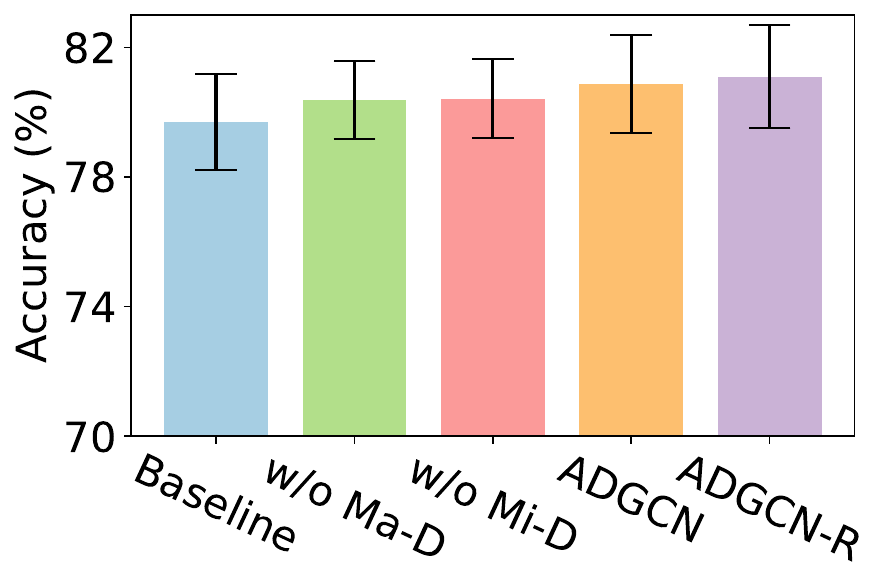}
		%\caption{fig1}
	}%
	\subfigure[Citeseer]{
		\centering
		\includegraphics[width=1.3in]{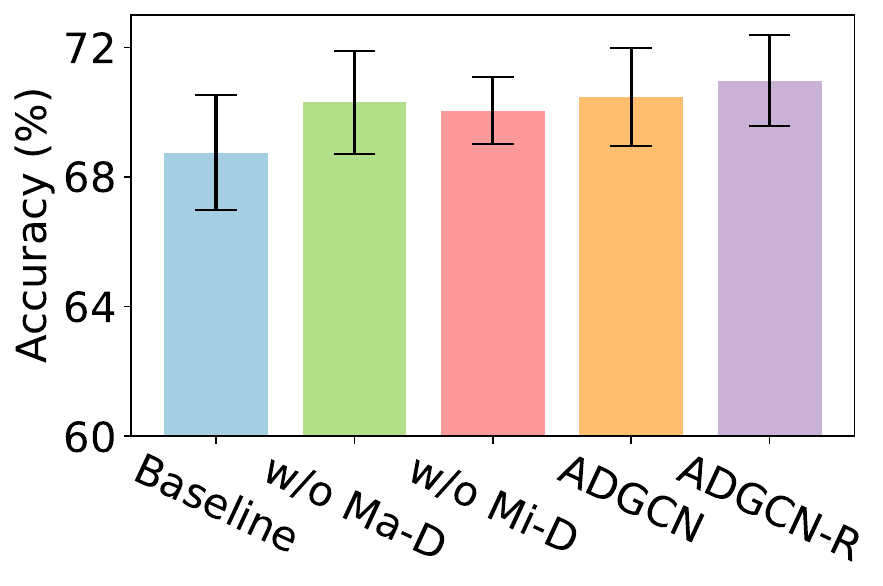}
		%\caption{fig1}
	}%
	\vspace{-6pt}
	\caption{Effectiveness evaluation of Micro-D, Macro-D, and graph refinement, where w/o Micro-D and w/o Marco-D represents ADGCN without micro disentanglement and ADGCN without macro disentanglement, respectively.} %We use the results of semi-supervised node classification to verify the effectiveness of each part.}
        \vspace{-15pt}
	\label{ablation_study}

\end{figure}

\begin{figure*}[t]
	\vspace{-0.6cm}
         \subfigcapskip=-6pt
	\subfigure[1st component space]{
        
		\centering
		\includegraphics[width=1.3in]{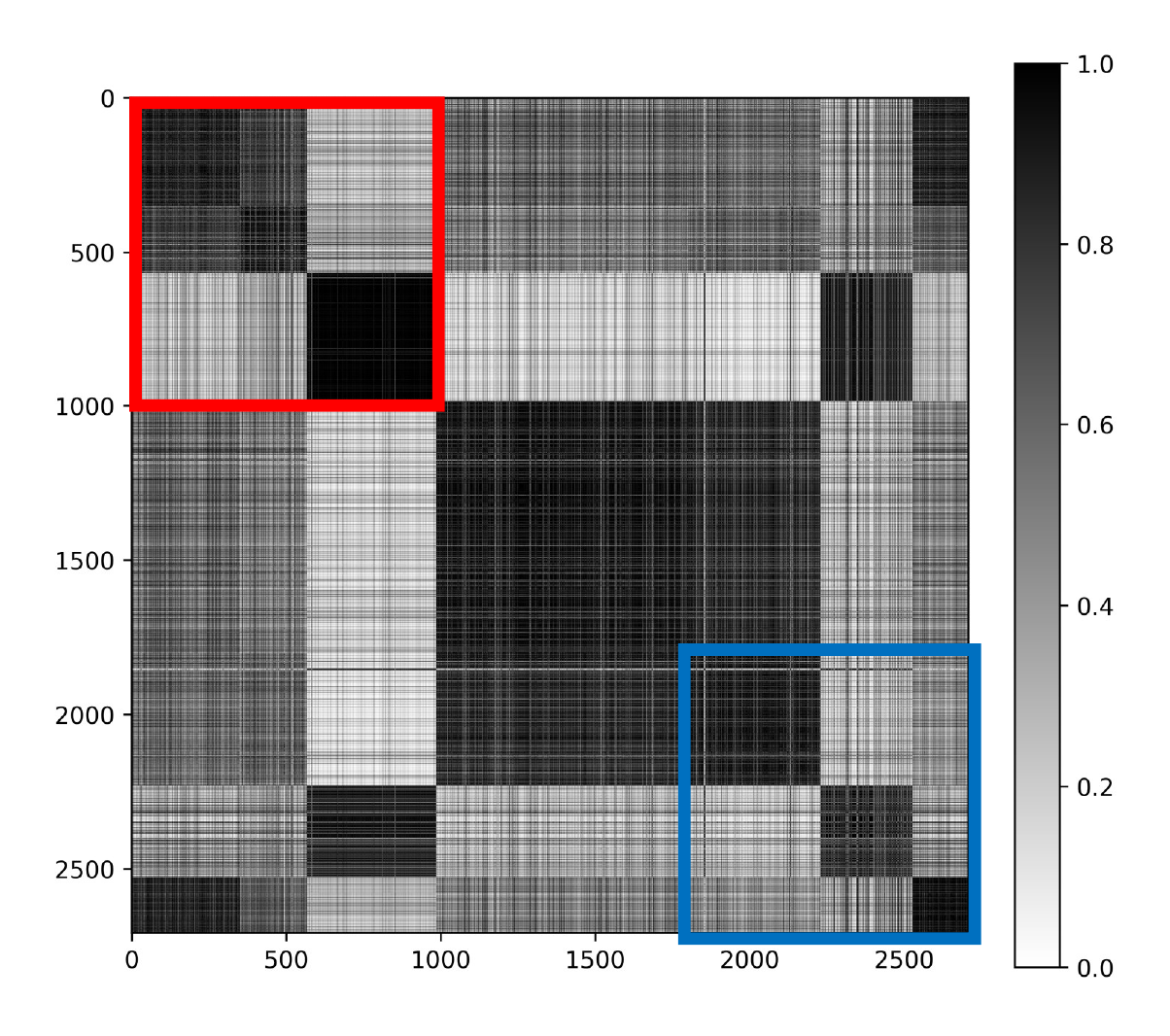}
		%\caption{fig1}
	}%
	\centering
	\subfigure[2nd component space]{
		\centering
		\includegraphics[width=1.3in]{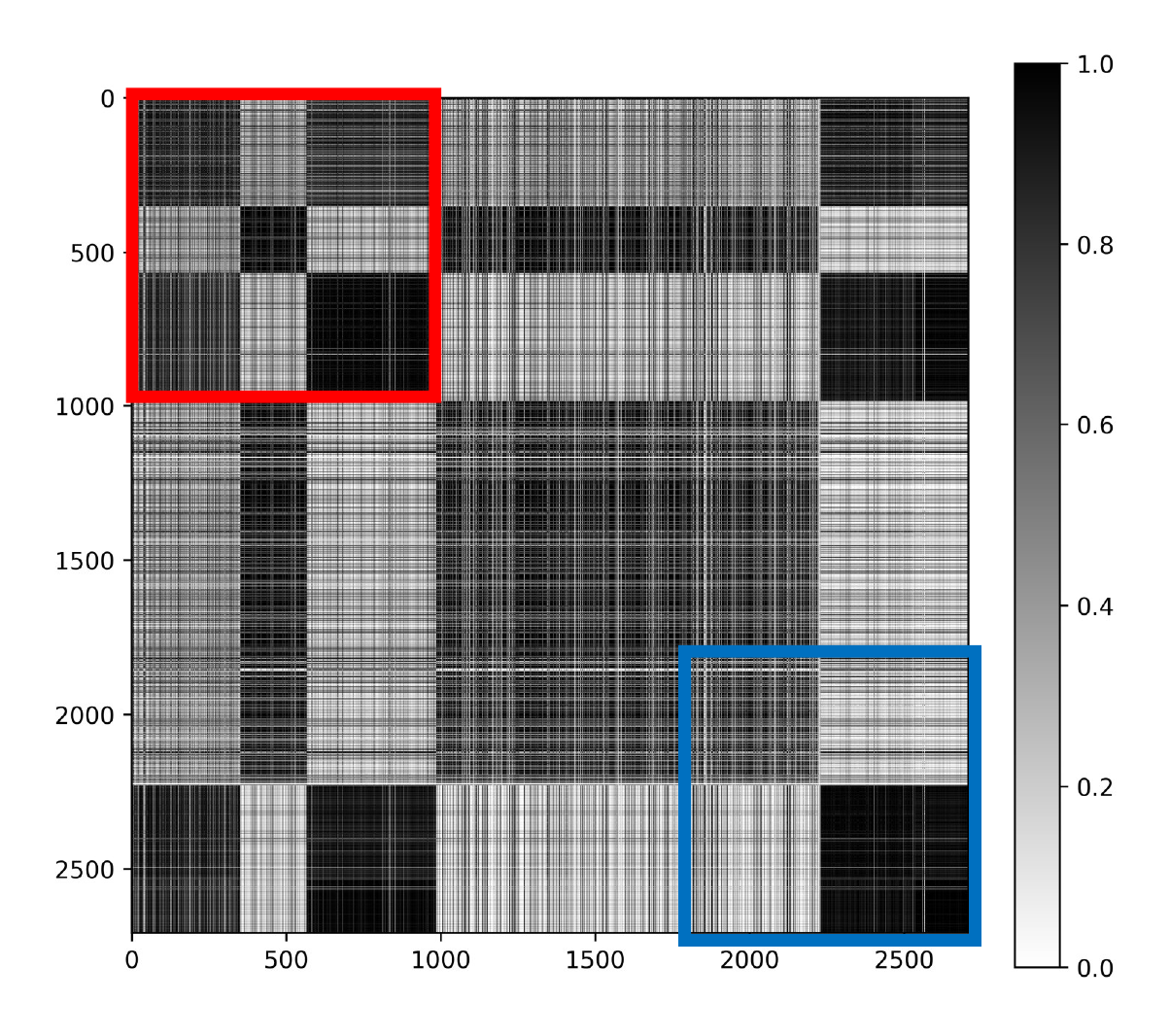}
		%\caption{fig2}
	}%
	%\\
	\subfigure[3rd component space]{
		\centering
		\includegraphics[width=1.3in]{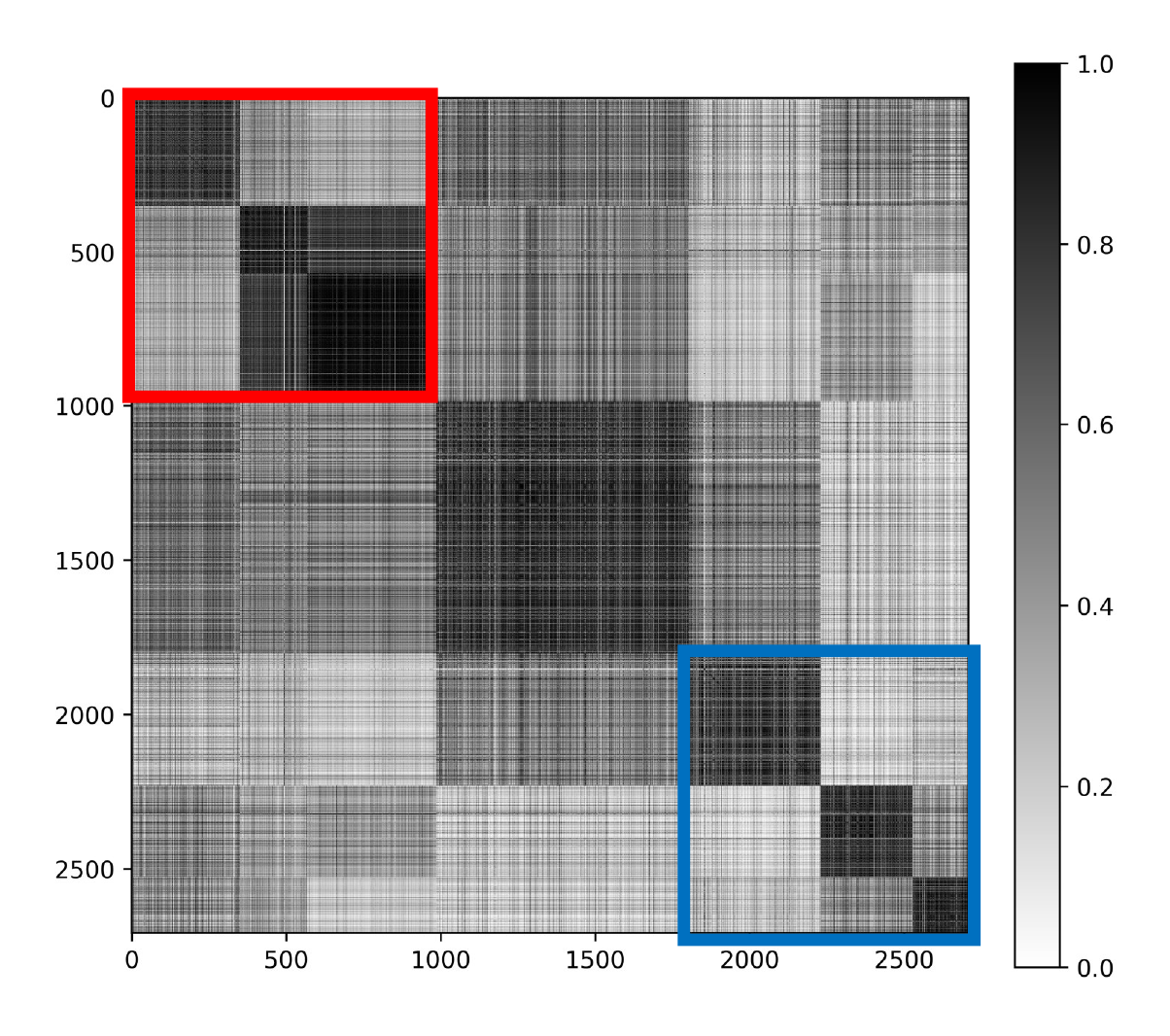}
		%\caption{fig2}
	}%
	\subfigure[4th component space]{
	\centering
	\includegraphics[width=1.3in]{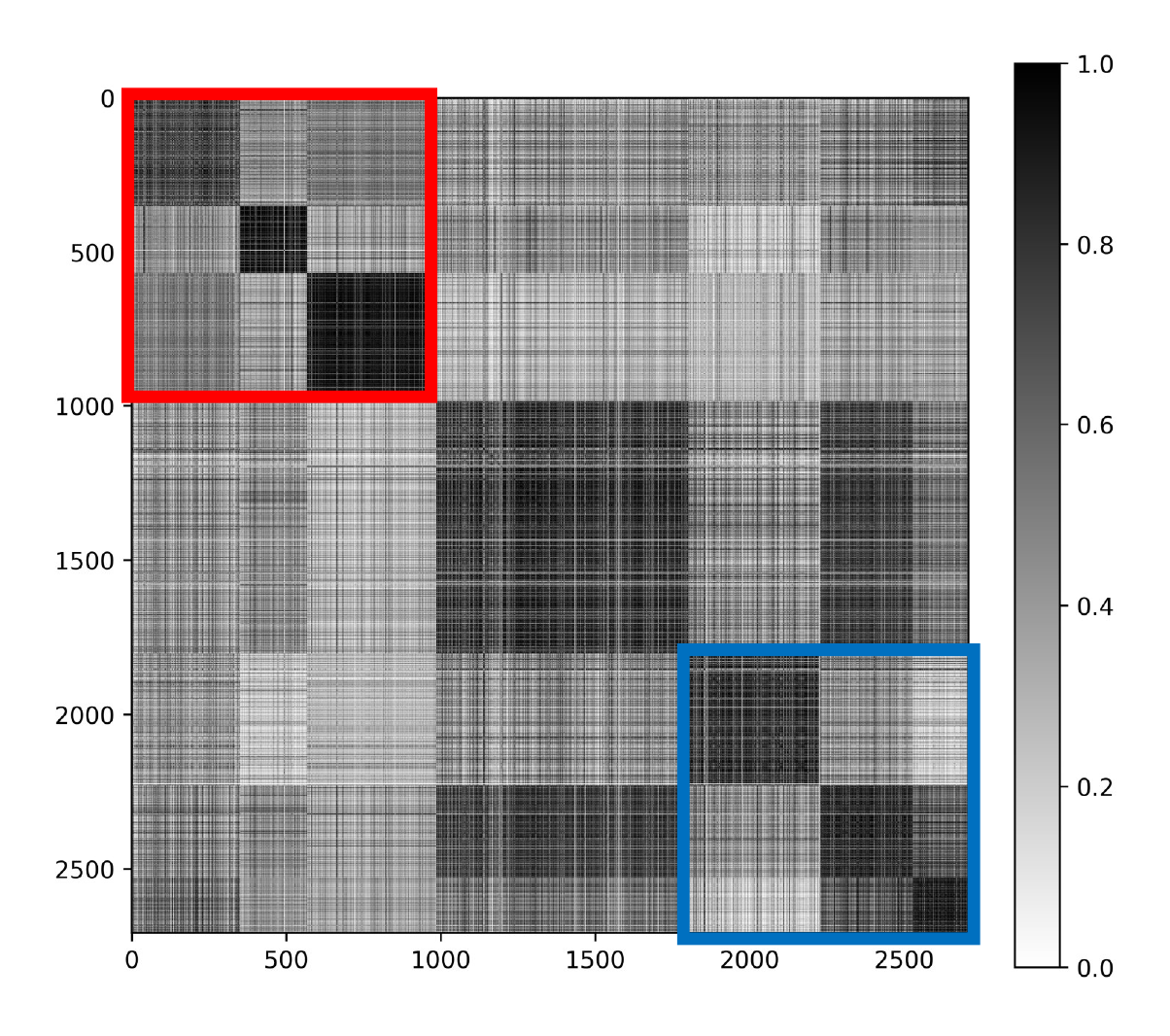}
	%\caption{fig2}
}%
	\subfigure[5th component space]{
	\centering
	\includegraphics[width=1.3in]{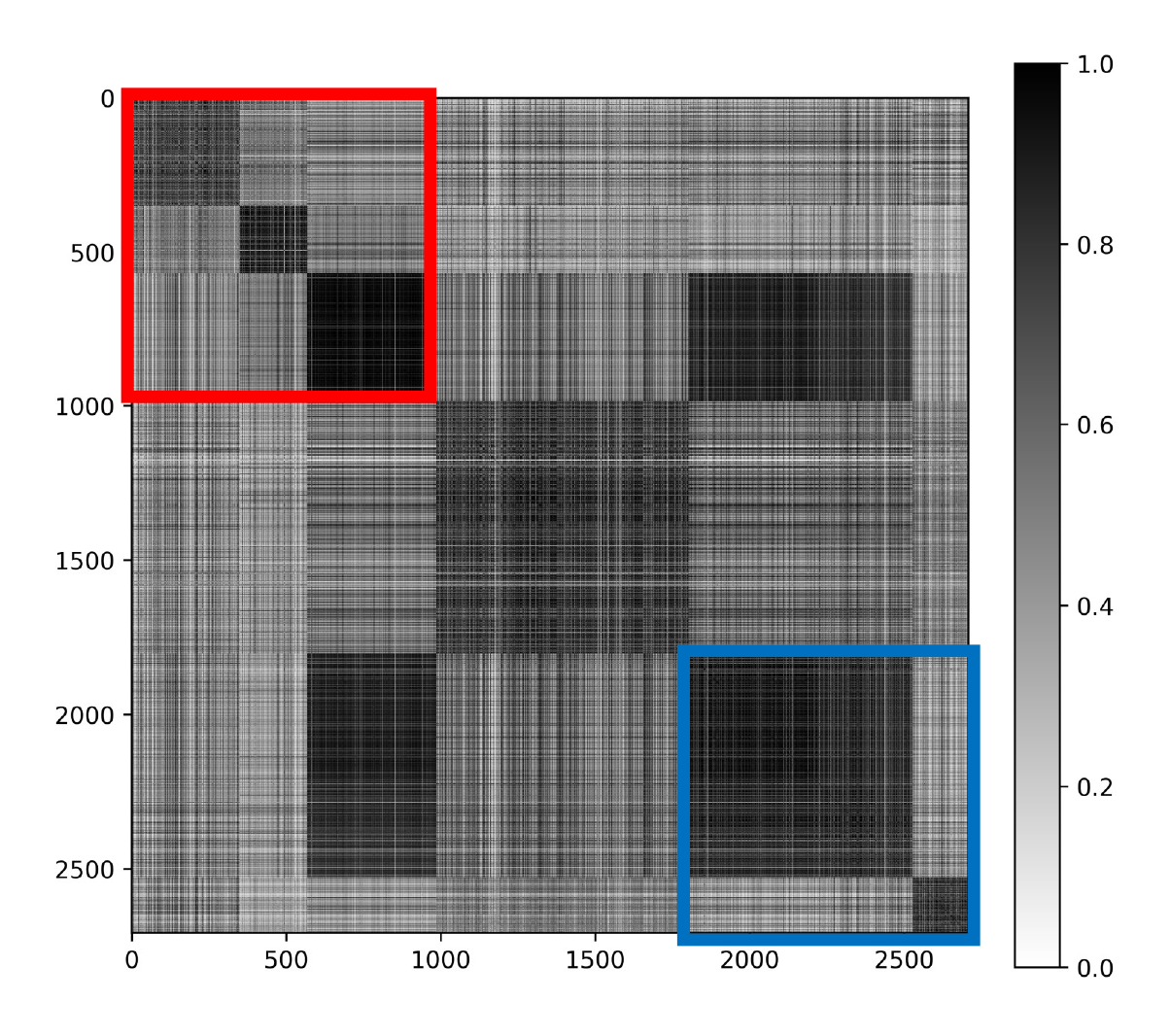}
	%\caption{fig2}
}%
	\centering
        \vspace{-0.2cm}
	\caption{Visualization of the similarity matrix of node representations in different component spaces on the Cora dataset. We use the red and blue box markers to show the entangled relationship of nodes in different component spaces.}
	\centering
	\label{component visualization 2}
	\vspace{-0.2cm}
\end{figure*}

\begin{figure}[t]
	\centering
    \subfigcapskip=-5pt
    \subfigure[Original node feature $\mathbf{X}$]{
		\centering
		\includegraphics[width=1.3in]{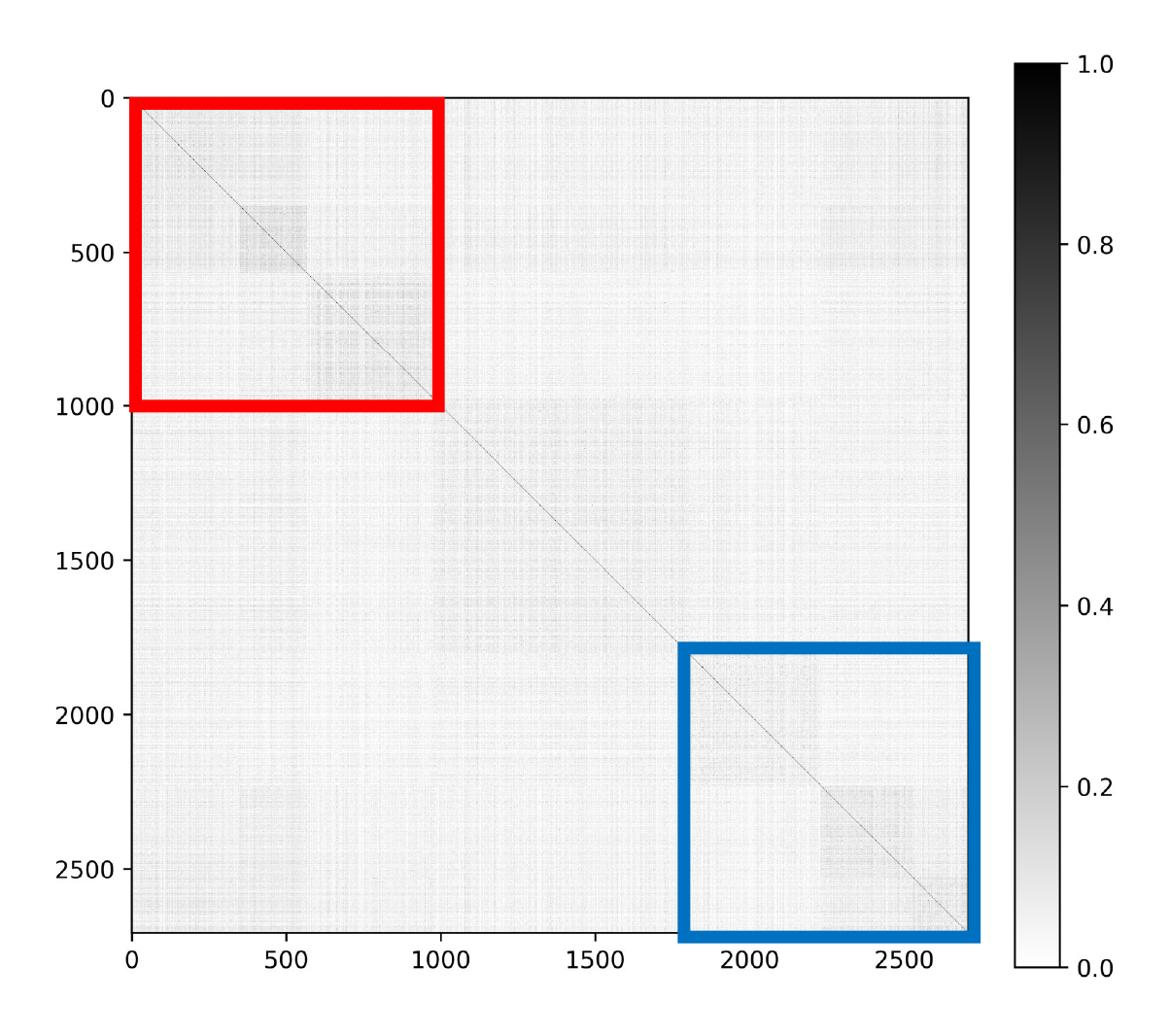}
		%\caption{fig1}
	}%
	\subfigure[Learned representation $\mathbf{H}$]{
		\centering
		\includegraphics[width=1.3in]{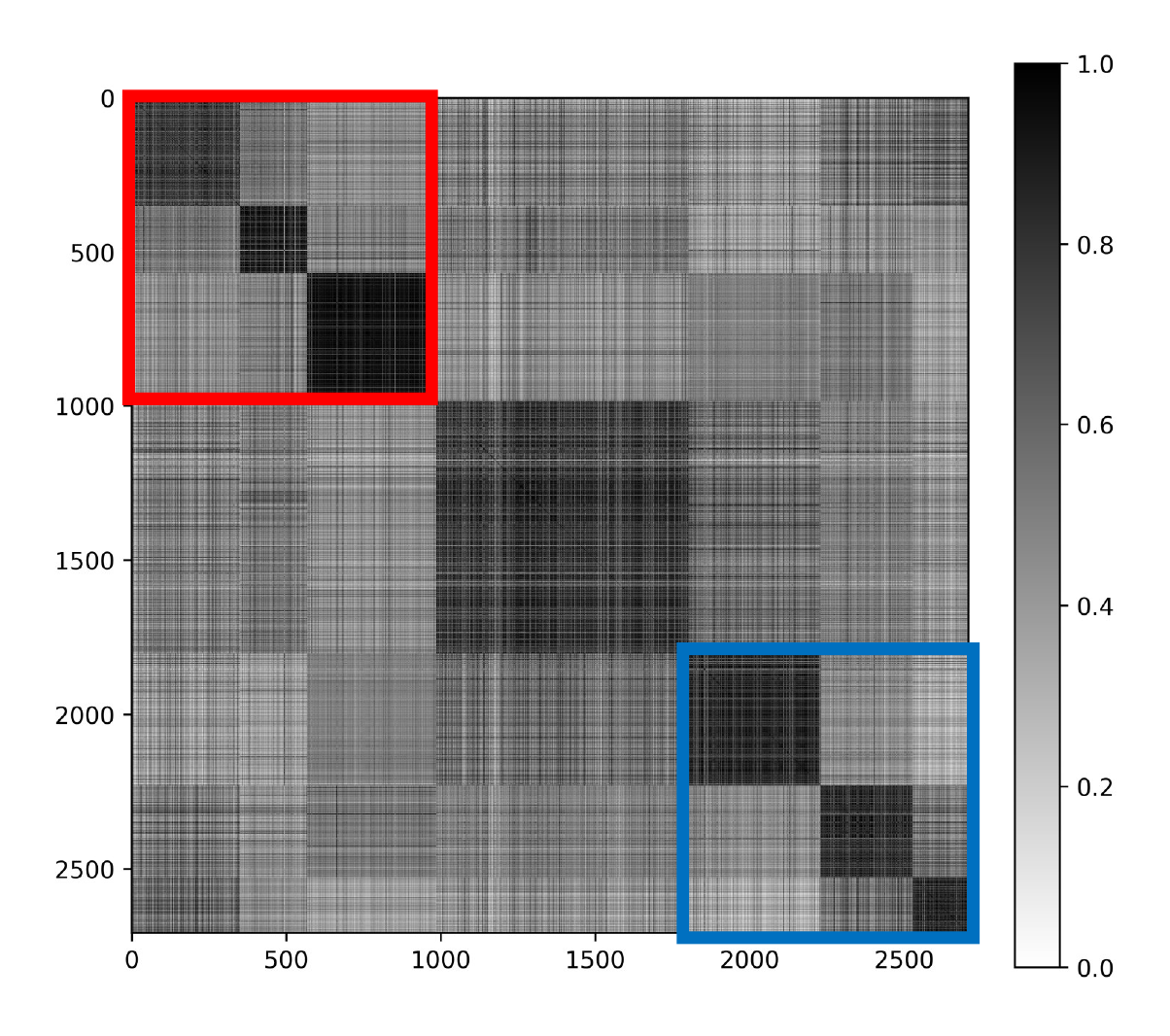}
		%\caption{fig1}
	}%
	\vspace{-0.1cm}
	\caption{Visualization of (a) the similarity matrix of node feature $\mathbf{X}$, (b) the similarity matrix of learned node representations on the Cora dataset. }
	\vspace{-0.cm}
	\label{component visualization}
\end{figure}
\vspace{-0.3cm}
\subsection{Clustering Analysis}

For the node clustering task, we use the K-means~\cite{Kmeans} to cluster the learned representations as in~\cite{IPGDN}. The setting of the three citation datasets is the same as in the semi-supervised node classification task with the fixed split.
As shown in Table~\ref{node_clustering}, the performances of graph disentanglement methods show a significant advantage over other compared methods. Moreover, our ADGCN achieves the best performance compared to DisenGCN and IPGDN.
It means that the proposed macro and micro disentanglement are greatly beneficial to exploring the inherent similarity between nodes. Meanwhile, the excellent performance of these graph disentanglement-based methods also demonstrates that graph disentanglement can indeed provide benefits in terms of graph clustering.

\begin{table*}[t]
  \begin{minipage}[p]{0.7\textwidth} 
    \caption{GPU memory cost and average training time per epoch.}
    \centering
    \vspace{-5pt}
\scriptsize
\setlength\tabcolsep{1.25pt}
\renewcommand{\arraystretch}{1.1}

\begin{tabular}{cccccc|ccccc} 
\hline
\multirow{2}{*}{Dataset} & \multicolumn{5}{c|}{Max GPU memory (MB)}        & \multicolumn{5}{c}{Average training time per epoch (ms)}  \\ 
\cline{2-11}
                         & DisenGCN & FactorGCN & DisGNN & ADGCN & ADGCN-R & DisenGCN & FactorGCN & DisGNN & ADGCN & ADGCN-R           \\ 
\hline
Cora                     & 1847     & 1771         & 9239      & 1753  & 1753    & 201.6    & 227.3    & 492.1      & 190.5 & 212.0             \\
Citeseer                 & 2179     & 1319         & 15471      & 2267  & 2267    & 214.8    & 475.0     & 730.7      & 241.3 & 337.1             \\
Pubmed                   & 4495     & 1549         & 19695      & 3887  & 3887    & 698.6    & 818.9    & 10673.2      & 821.4 & 942.4             \\
CS                       & 2647     & 2473         & OOM      & 2365  & 2365    & 244.2    & 797.1   & OOM      & 220.5 & 411.2             \\
Computers                & 2275     & 1589         & OOM      & 2505  & 2505    & 452.4    & 1091.3      & OOM      & 462.8 & 823.7             \\
Photo                    & 2031     & 1411         & OOM      & 1739  & 1739    & 191.0    & 721.0      & OOM      & 169.7 & 465.6             \\
\hline
\end{tabular}
\label{cost}
  \end{minipage}
\begin{minipage}[p]{0.3\textwidth}
\caption{The time cost of each part of ADGCN.}
\vspace{-5pt}
\centering
\scriptsize
\setlength\tabcolsep{1.5pt}
\renewcommand{\arraystretch}{1.1}

\begin{tabular}{cccc}
\hline
\multirow{2}{*}{Dataset} & \multicolumn{3}{c}{Average training time per epoch (ms)} \\ \cline{2-4} 
                         & ADGCN w/o Micro-D          & ADGCN          & ADGCN-R         \\ \hline
Cora                     & 166.5                   & 190.5           & 212.0           \\
Citeseer                 & 214.9                   & 241.3           & 337.1           \\
Pubmed                   & 693.6                   & 821.4           & 942.4           \\
CS                       & 191.4                   & 220.5           & 411.2           \\
Computers                & 422.9                   & 462.8           & 823.7           \\
Photo                    & 134.3                   & 169.7           & 465.6           \\ \hline
\end{tabular}

\label{each_part_cost}
\end{minipage}
\vspace{-0.6cm}
\end{table*}

\begin{table}[t]\scriptsize
	\centering
	\caption{The node classification accuracies of different graphs.}
	\begin{tabular}{cccc}
		\hline
		\multirow{2}{*}{Graph} & \multicolumn{3}{c}{Datasets}                                                        \\ \cline{2-4}
		                    & Cora             & Citeseer                   & Pubmed \\ \hline
		$\mathbf{A}$      	& 68.0$\pm$1.44          & 45.3$\pm$0.91          & 63.0$\pm$1.92            \\ \hline
		$\mathbf{A}_{k-NN}$ & 74.9$\pm$1.95          & 57.5$\pm$1.00          & 69.4$\pm$1.52            \\ \hline
		$\mathbf{A}^{(T)}$  & \textbf{77.3$\pm$1.38} & \textbf{63.5$\pm$1.29} & \textbf{72.6$\pm$1.55}   \\ \hline
	\end{tabular}
	
	\label{three_graph}
 	\vspace{-0.3cm}
\end{table}

\subsection{Ablation Study}
We conduct an ablation study through the semi-supervised node classification on Cora and Citeseer datasets. As shown in Fig.\ref{ablation_study}, the proposed macro disentanglement, micro disentanglement, and diversity-preserving graph refinement are equally important to learning latent node representation. Here, the baseline model refers to the GCN with multi-component. Specifically, both ADGCN w/o Macro-D and ADGCN w/o Micro-D achieve better performance than the baseline. More importantly, the performance of the combination of Macro-D and Micro-D (i.e., ADGCN) verifies that both Macro-D and Micro-D are mutually reinforcing. Besides, it can be seen that the adoption of the progressive graph refinement is also helpful for further performance improvement.

\textcolor{black}{To verify the effectiveness of the progressive graph refinement further, the label propagation~\cite{lp} is applied to evaluate the ability of the refined adjacency matrix to characterize node relationships. The performance on the two datasets is shown in Table \ref{three_graph}, where $\mathbf{A}_{k-NN}$ and $\mathbf{A}^{(T)}$ denote the adjacency matrix constructed by the learned representation $H^{(T)}$ and the final adjacency matrix refined through progressive graph refinement, respectively. As we can see that $\mathbf{A}^{(T)}$ performs better than both $\mathbf{A}$ and $\mathbf{A}_{k-NN}$ by about $9.3\%$ and $2.4\%$, respectively, on Cora dataset. The results show that the graph refinement strategy can effectively boost the node representation and the graph structure collaboratively.}

\textcolor{revise_color}{We also evaluate the influence of the two regularization terms of conditional adversarial learning on our ADGCN, i.e., $\mathcal{L}_{adv}$ and $\mathcal{L}_{cls}$. 
As shown in Table~\ref{exp::adversarial_learning}, both $\mathcal{L}_{adv}$ and $\mathcal{L}_{cls}$ have an positive role to the performance of ADGCN. 
Concretely, the performance gain is just modest when only the $\mathcal{L}_{cls}$ is adopted, and different forms of the adversarial objective $\mathcal{L}_{adv}$ will bring various performance influences. }

\subsection{Analysis of Disentanglement}
To intuitively show the disentanglement achieved by ADGCN, we visualize the similarity matrix of the Cora dataset in both the original feature and the learned representation spaces.
In contrast to the original feature in Fig.~\ref{component visualization}(a), the similarity matrix based on the learned representation has shown an explicit block effect as we can see from  
Fig.~\ref{component visualization}(b)
. It indicates that the representation learned by our model is capable of effectively capturing intra-class similarity and inter-class differences.

\begin{figure}[t]
	\centering
        \subfigcapskip=-3pt
	\subfigure[DisenGCN]{
		\centering
		\includegraphics[width=1.in]{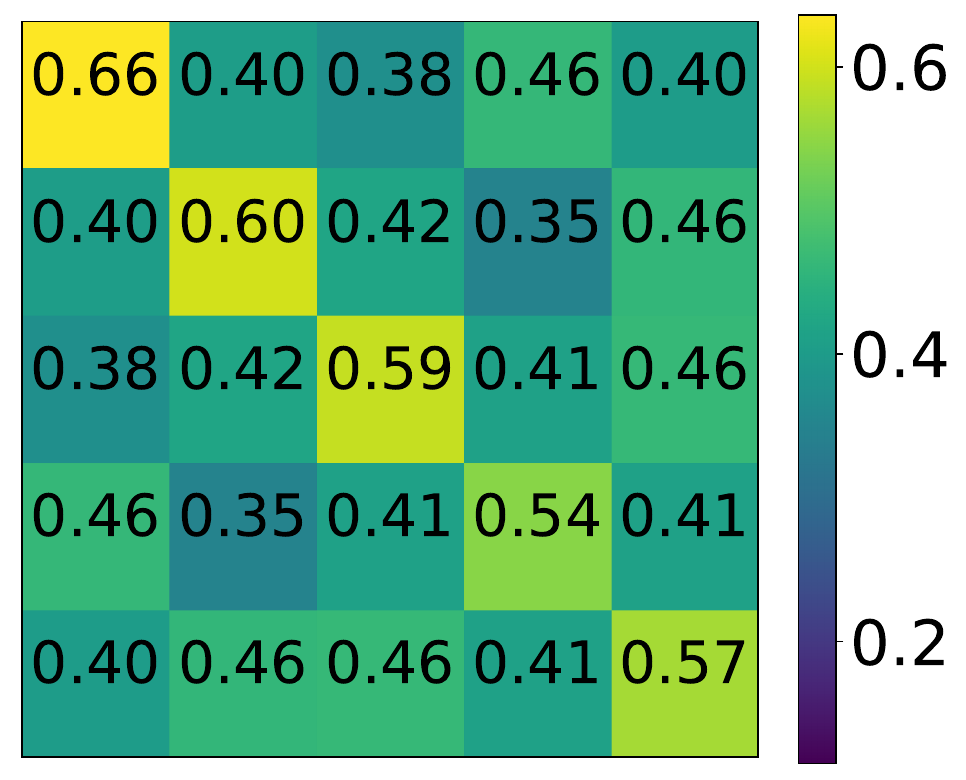}
		%\caption{fig1}
	}%
		\subfigure[w/o Macro-D]{
		\centering
		\includegraphics[width=1.in]{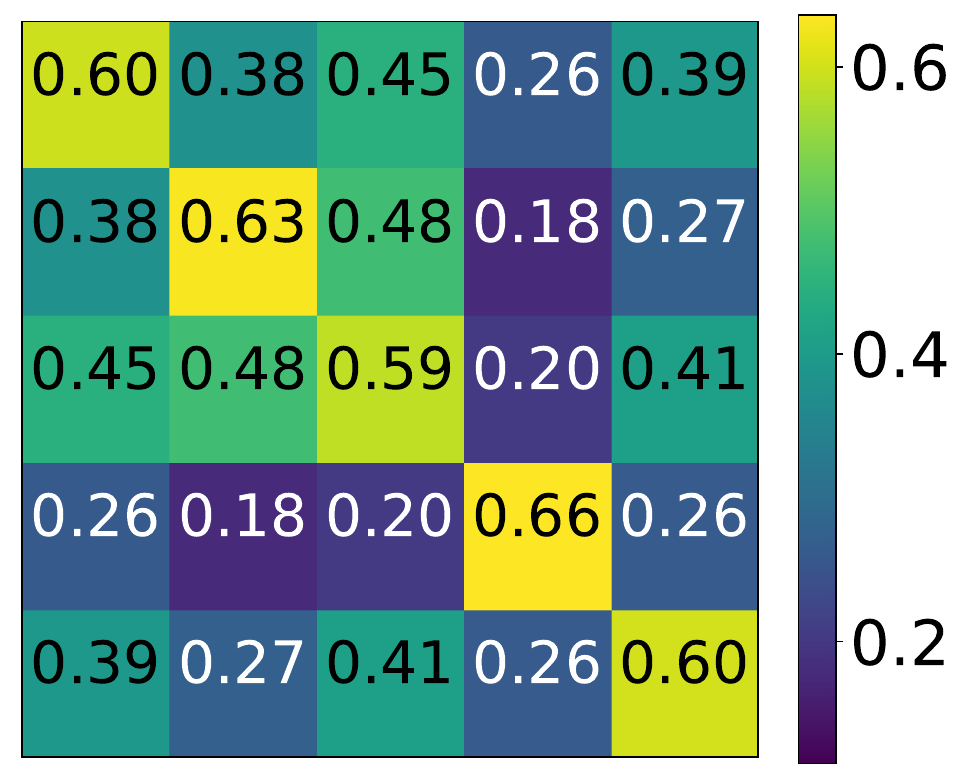}
		%\caption{fig1}
	}
	\subfigure[ADGCN]{
		\centering
		\includegraphics[width=1.in]{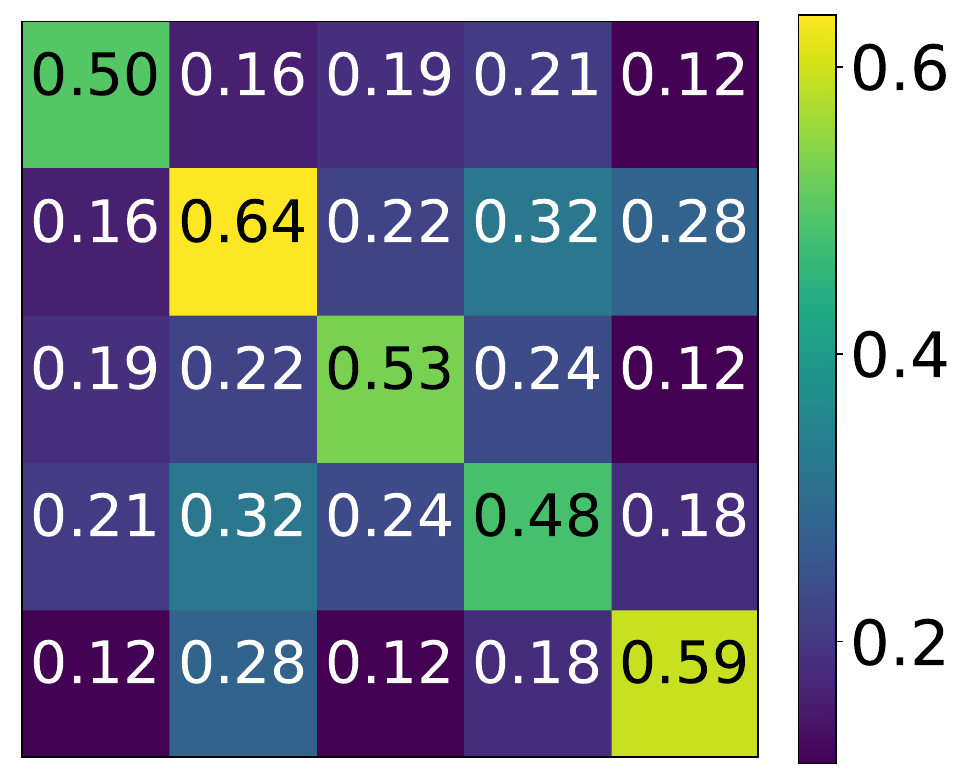}
		%\caption{fig1}
	}%
        \vspace{-0cm}
	\caption{\textcolor{black}{The component confusion matrix $C$ obtained by DisenGCN, ADGCN w/o Macro-D and ADGCN on Cora dataset.}}
 % Here, the group average similarity score is adopted to measure the correlation between different component distributions. The smaller the score, the weaker the correlation between the two component distributions.}
	\label{macro_disentanglement}
	\vspace{-0.2cm}
\end{figure}

More importantly, to reveal the intricate entanglement between different classes of nodes, we visualize the similarity matrices of the component level, thus showing the correlations between nodes in different component spaces. 
From Fig.~\ref{component visualization 2}, it can be observed that each component does reflect some entangled information among different classes at a certain level, and the complex entanglements between different two classes can be well unveiled by the different combinations of components.
Taking the first three classes (marked in red box) in Fig.~\ref{component visualization 2} as an example, class 1 and class 2 are significantly entangled but both can be well distinguished from class 3 in the 1st component space. These three classes also exhibit different entanglement relationships in the 2nd and 3rd component spaces, respectively. Similarly, the three classes marked by the blue box also show various entanglement relations in different component spaces.

As shown in Fig.~\ref{macro_disentanglement}, we also present the component confusion matrix $C$ to illustrate the correlations among the distributions in different component spaces. Here, we use the group average similarity to compute $C$. Specifically, the correlation score between the $i$-th and $j$-th component distributions is given by $C_{ij} = \frac{1}{|V|^2}\sum_{u,v \in V}\cos(\mathbf{h}^i_u,\mathbf{h}^j_v )$. The smaller the score, the weaker the correlation between the two component distributions. Obviously, the representations in different component subspaces learned by our ADGCN take weaker correlations compared to DisenGCN~\cite{DisenGCN}. 
\textcolor{black}{In particular, as we can find from  Fig.~\ref{macro_disentanglement}(b), ADGCN only with micro-disentanglement achieves considerable disentanglement performance compared to DisenGCN. Meanwhile, on the basis of micro-disentanglement, the proposed macro-disentanglement by adversarial learning can further boost the separability between components as shown in Fig.~\ref{macro_disentanglement}(c). }
%The visual results mentioned above are consistent with the expected graph disentanglement to a certain extent, indicating that our ADGCN achieves disentangled graph representation learning from both micro-disentanglement and macro-disentanglement perspectives.
\vspace{-0.2cm}
\subsection{Analysis of Convergence and Efficiency}

\begin{figure}[t]
	\centering
    \subfigcapskip=-5pt
    \subfigure[Cora]{
		\centering
		\includegraphics[width=1.4in]{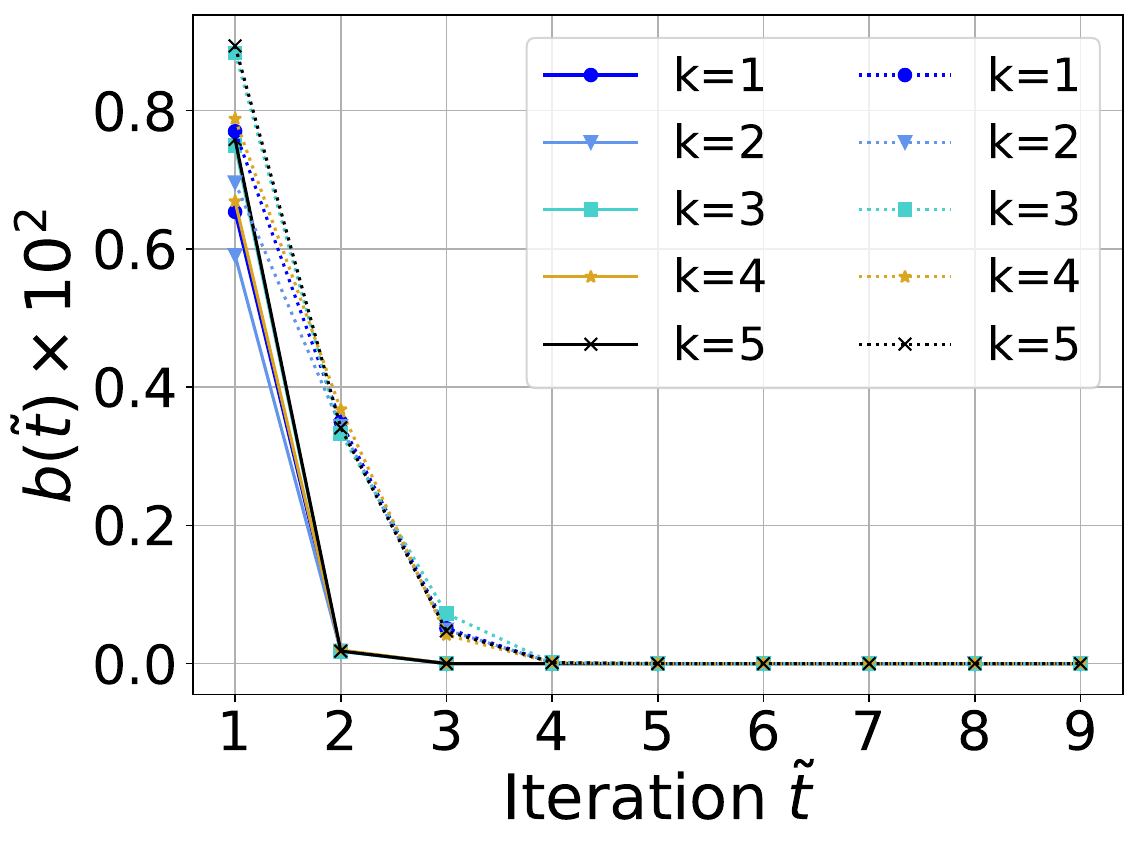}
		%\caption{fig1}
	}%
	\subfigure[Citeseer]{
		\centering
		\includegraphics[width=1.5in]{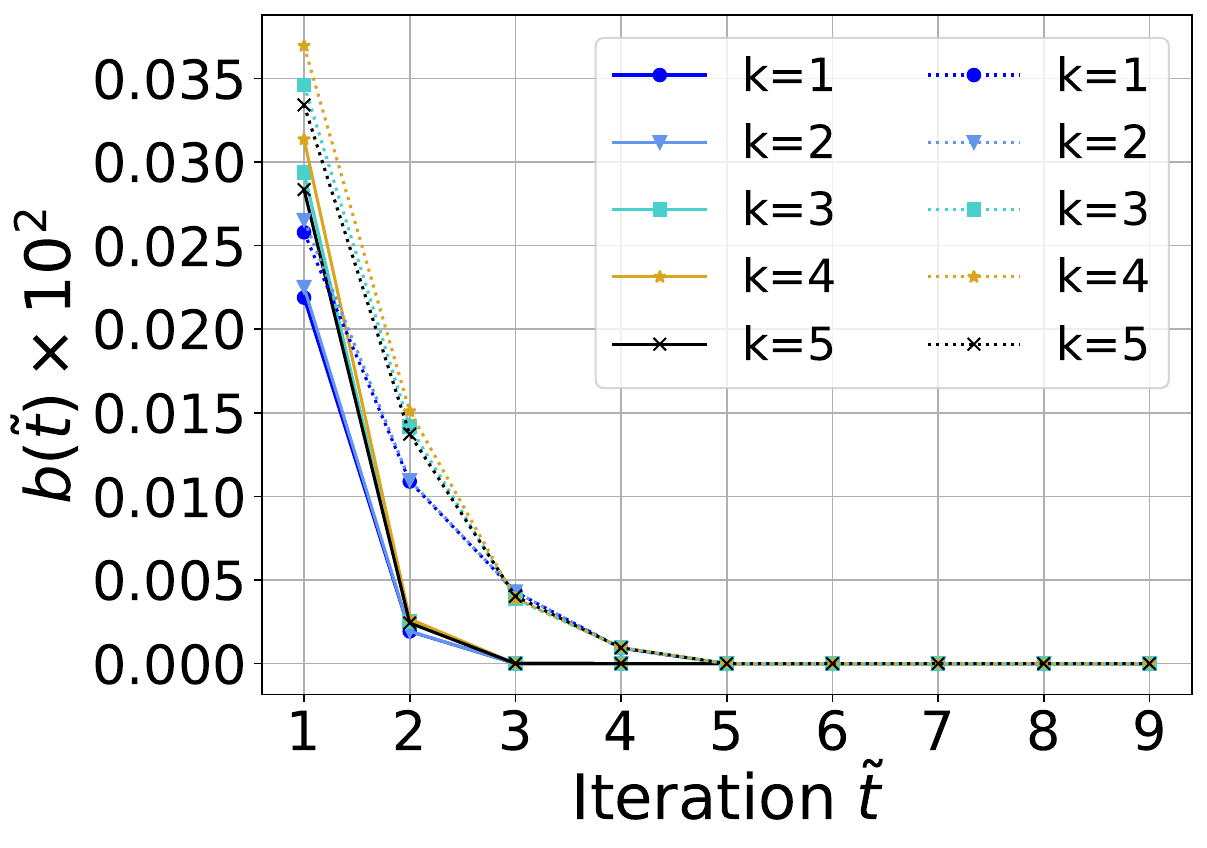}
		%\caption{fig1}
	}%
 \vspace{-0.2cm}
	\caption{The convergence analysis of the dynamic assignment on two datasets, where the solid lines ‘-` and the dotted lines '$\cdots$` denote the curves of ADGCN and ADGCN w/o $\mathcal{L}_{adv}$, respectively.}
	\label{convergence}
 \vspace{-0.5cm}
\end{figure}

% \begin{figure}[t]
% 	\centering
%     \subfigure[w/o Macro-D]{
% 		\centering
% 		\includegraphics[width=1.5in]{convergence/ADGCN_wo_macroD-loss_curve.png}
% 		%\caption{fig1}
% 	}%
% 	\subfigure[ADGCN]{
% 		\centering
% 		\includegraphics[width=1.5in]{convergence/ADGCN-loss_curve.png}
% 		%\caption{fig1}
% 	}%
% 	\caption{ The loss curves of model training on Cora dataset. }
% 	\label{loss_convergence}
% \end{figure}

\begin{figure*}[t]
	\centering
        \subfigcapskip=-5pt
	\subfigure[Cora-nettack]{
		\centering
		\includegraphics[height=1in]{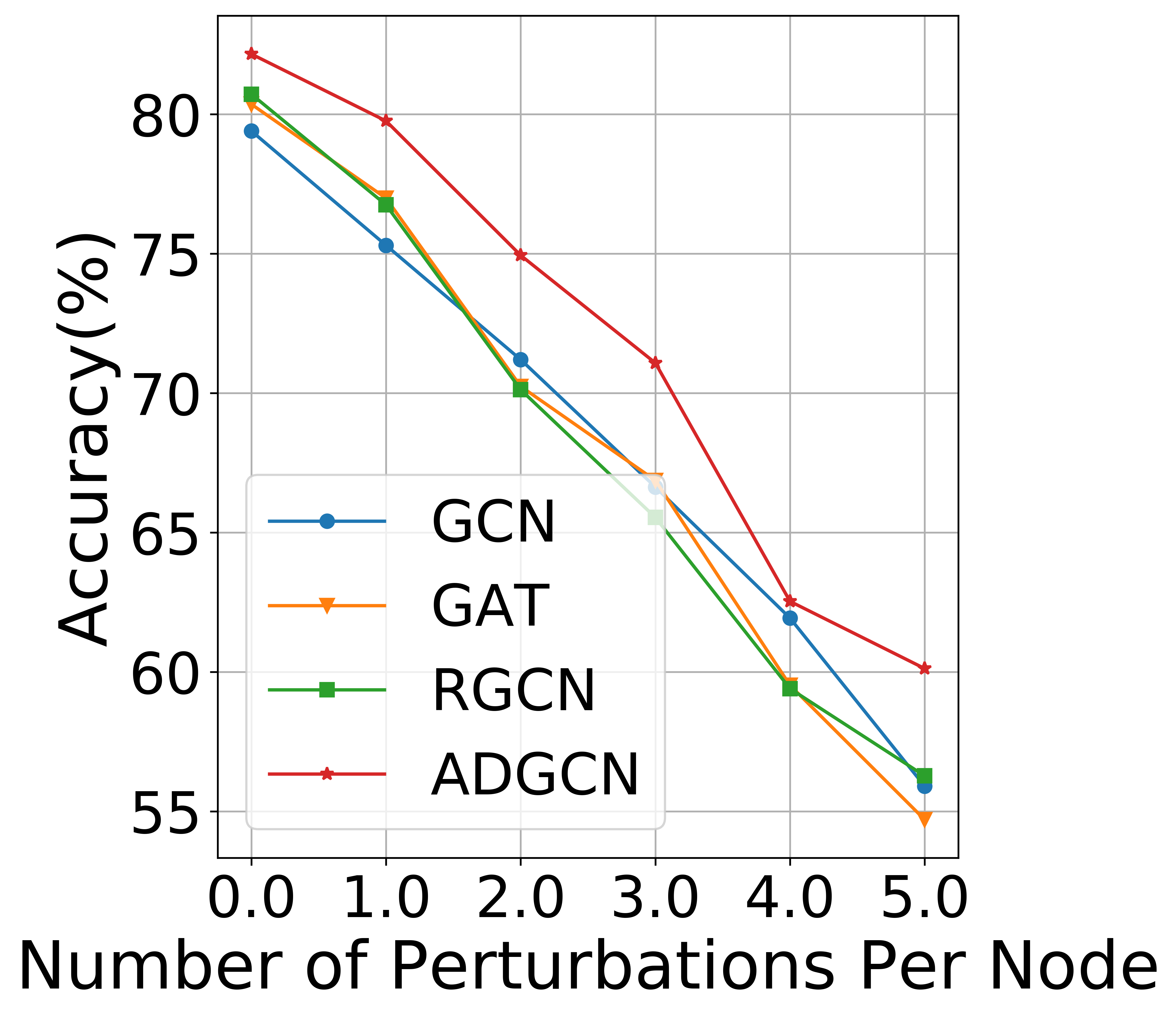}
		%\caption{fig1}
	}%
	\subfigure[Citeseer-nettack]{
		\centering
		\includegraphics[height=1in]{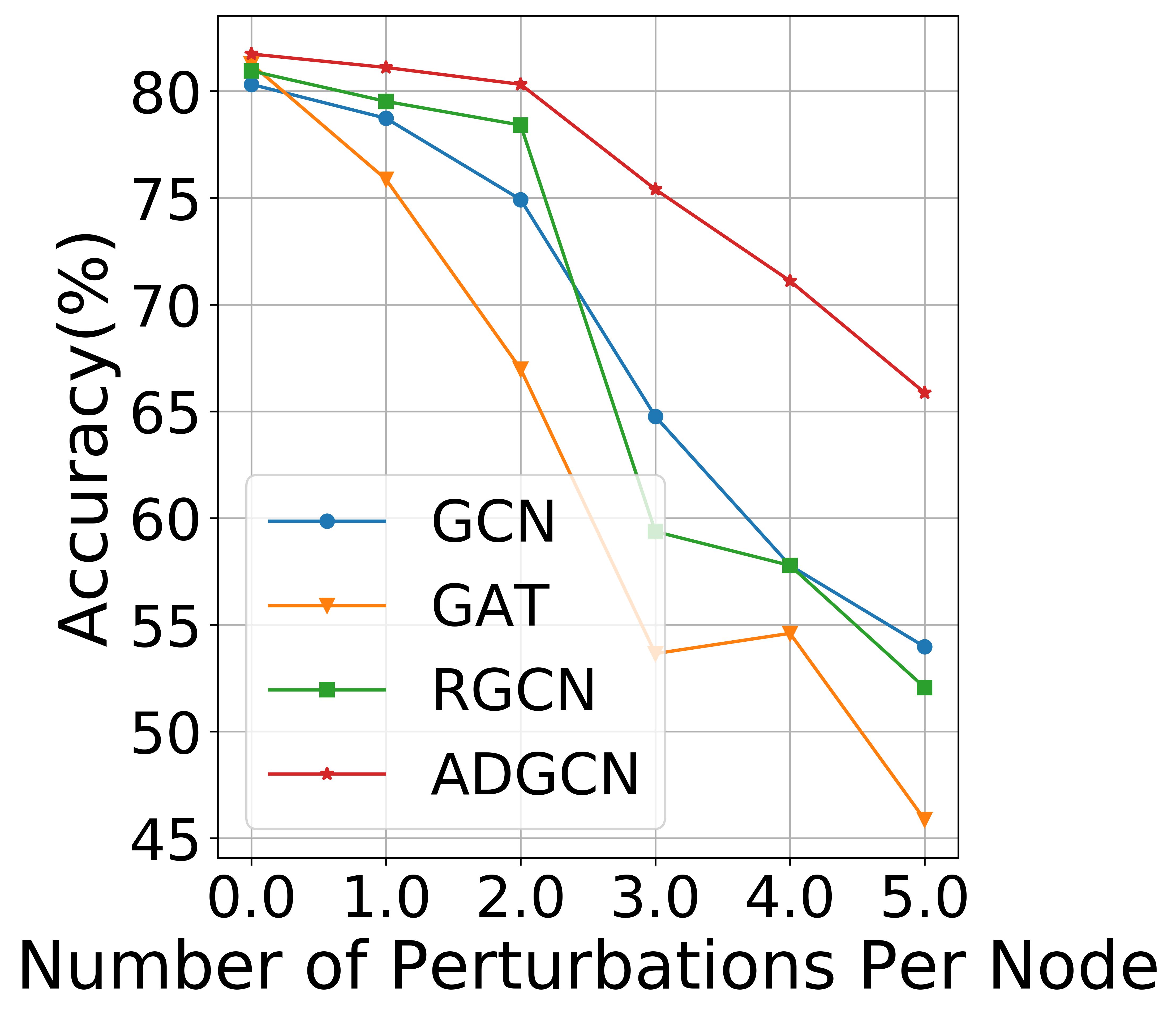}
		%\caption{fig1}
	}%
		\subfigure[Pubmed-nettack]{
		\centering
		\includegraphics[height=1in]{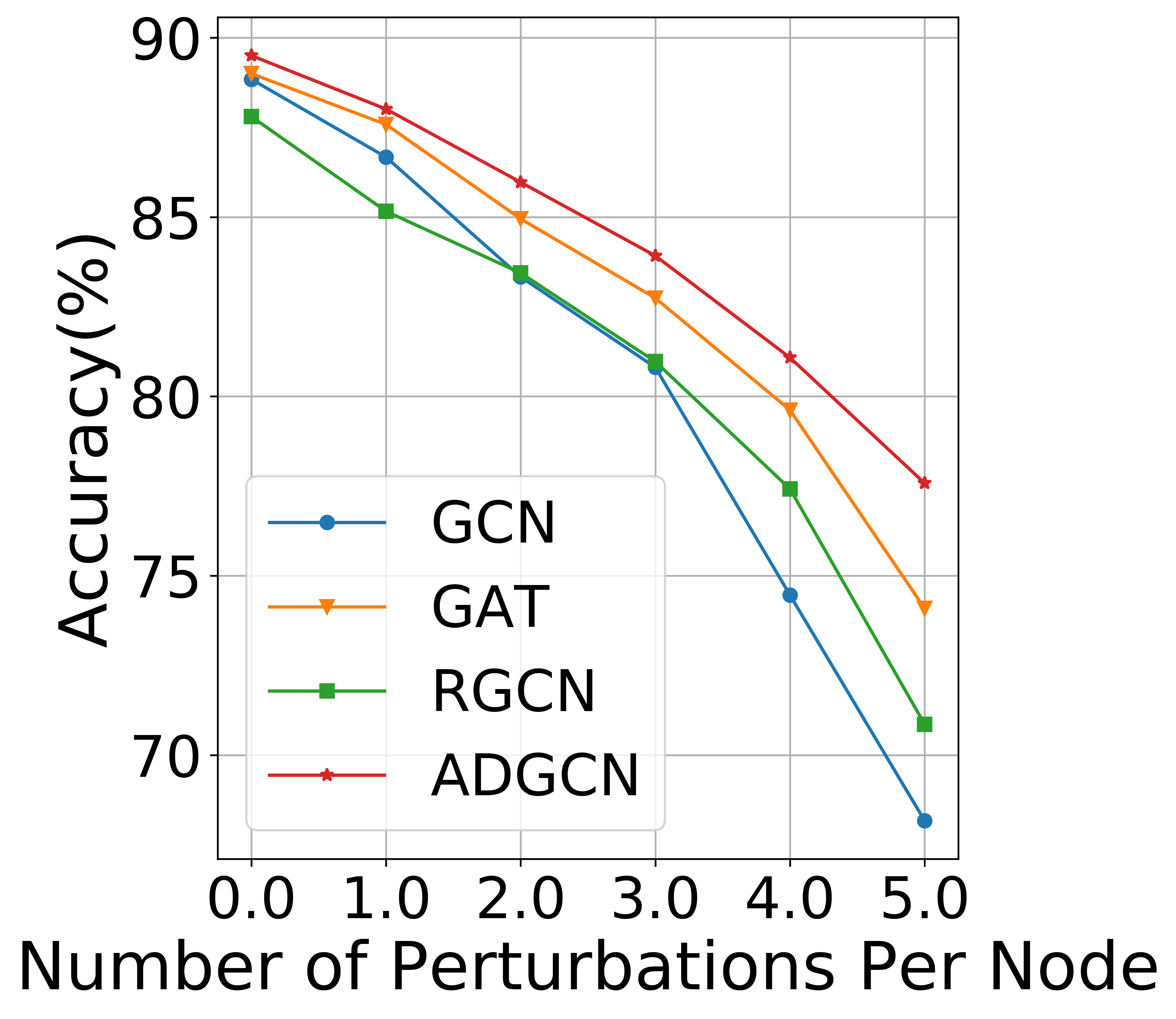}
		%\caption{fig1}
	}%
		\subfigure[Cora-metattack]{
		\centering
		\includegraphics[height=1in]{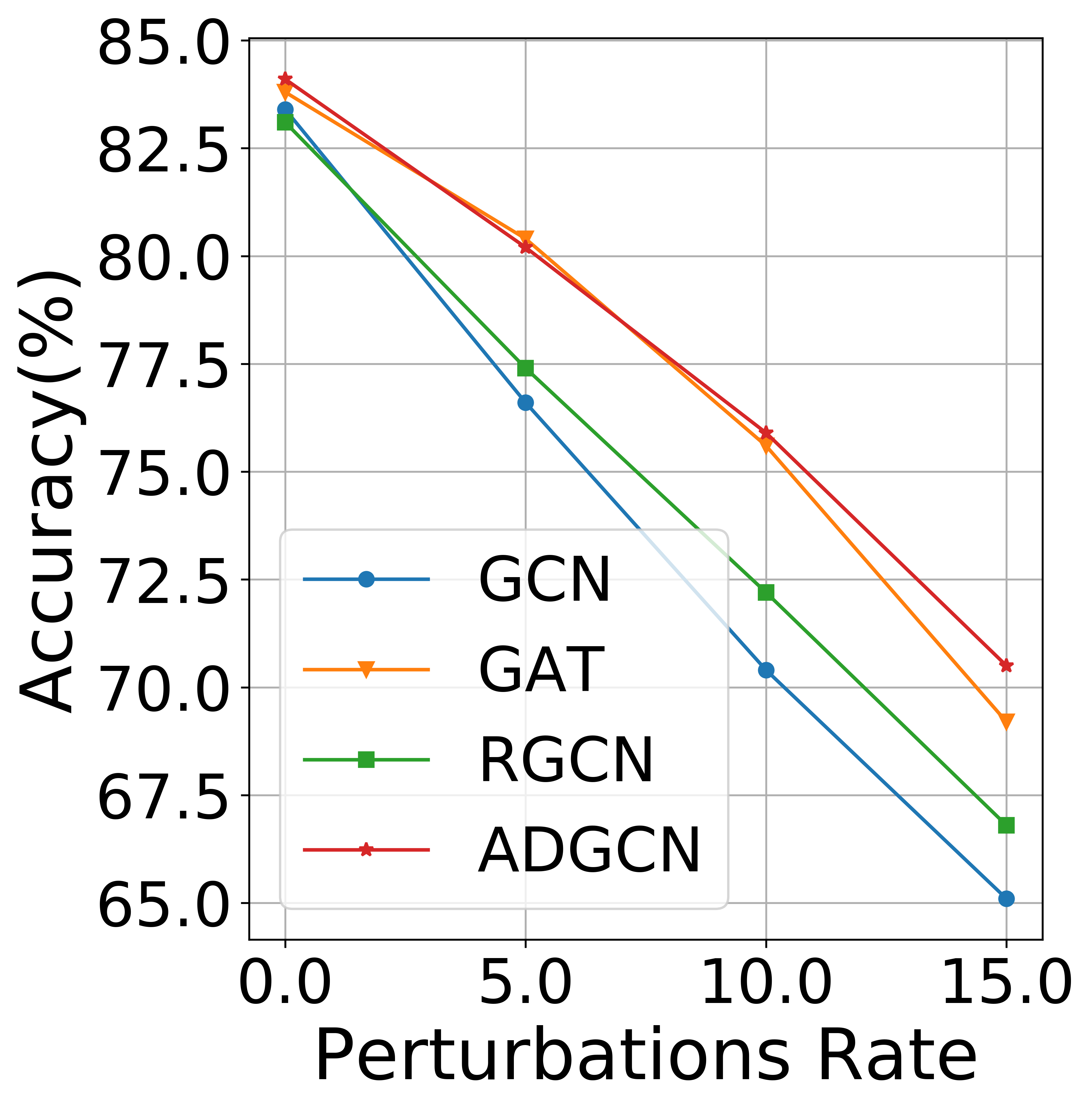}
		%\caption{fig1}
	}%
		\subfigure[Citeseer-metattack]{
		\centering
		\includegraphics[height=1in]{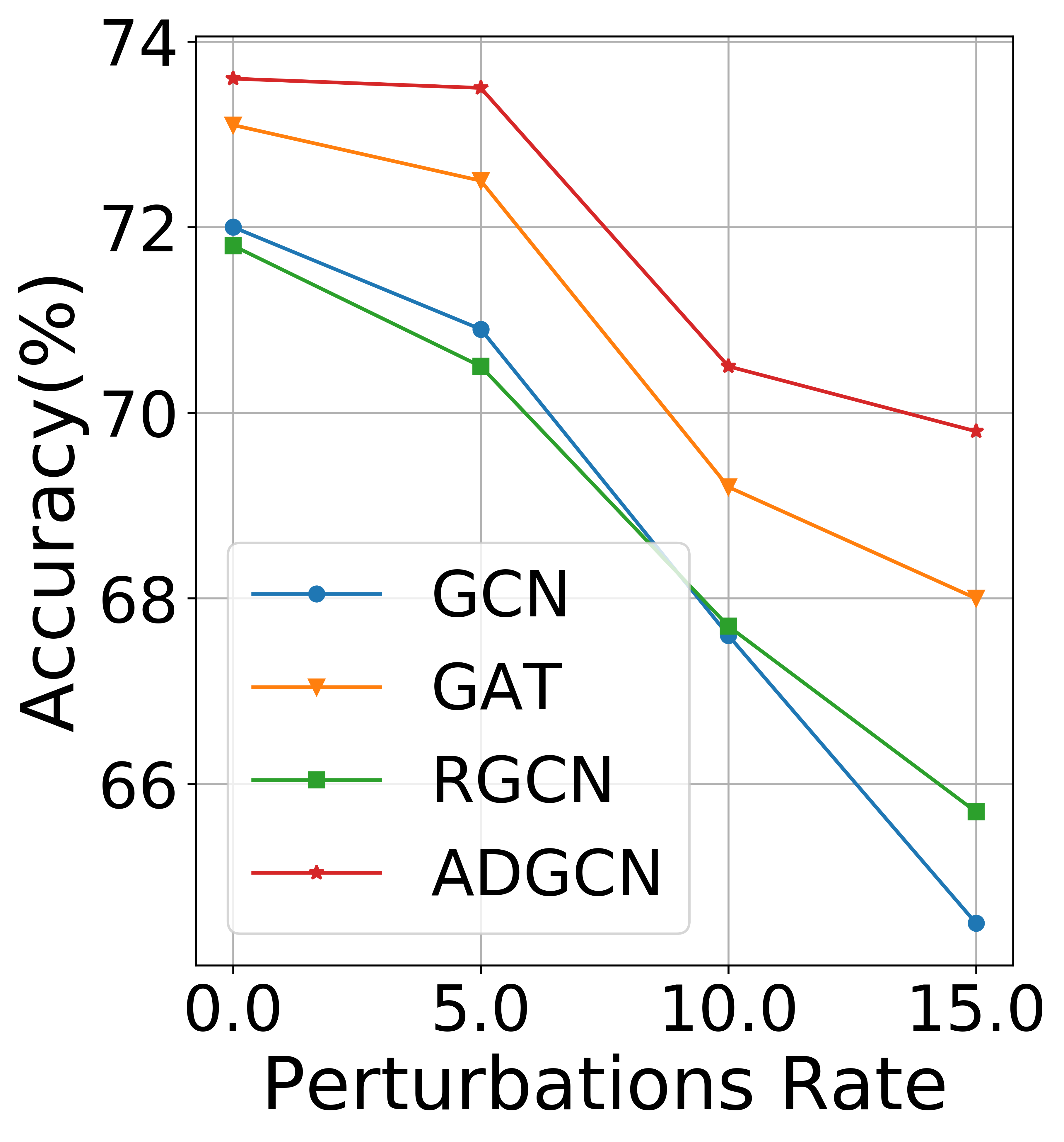}
		%\caption{fig1}
	}%
		\subfigure[Pubmed-metattack]{
		\centering
		\includegraphics[height=1in]{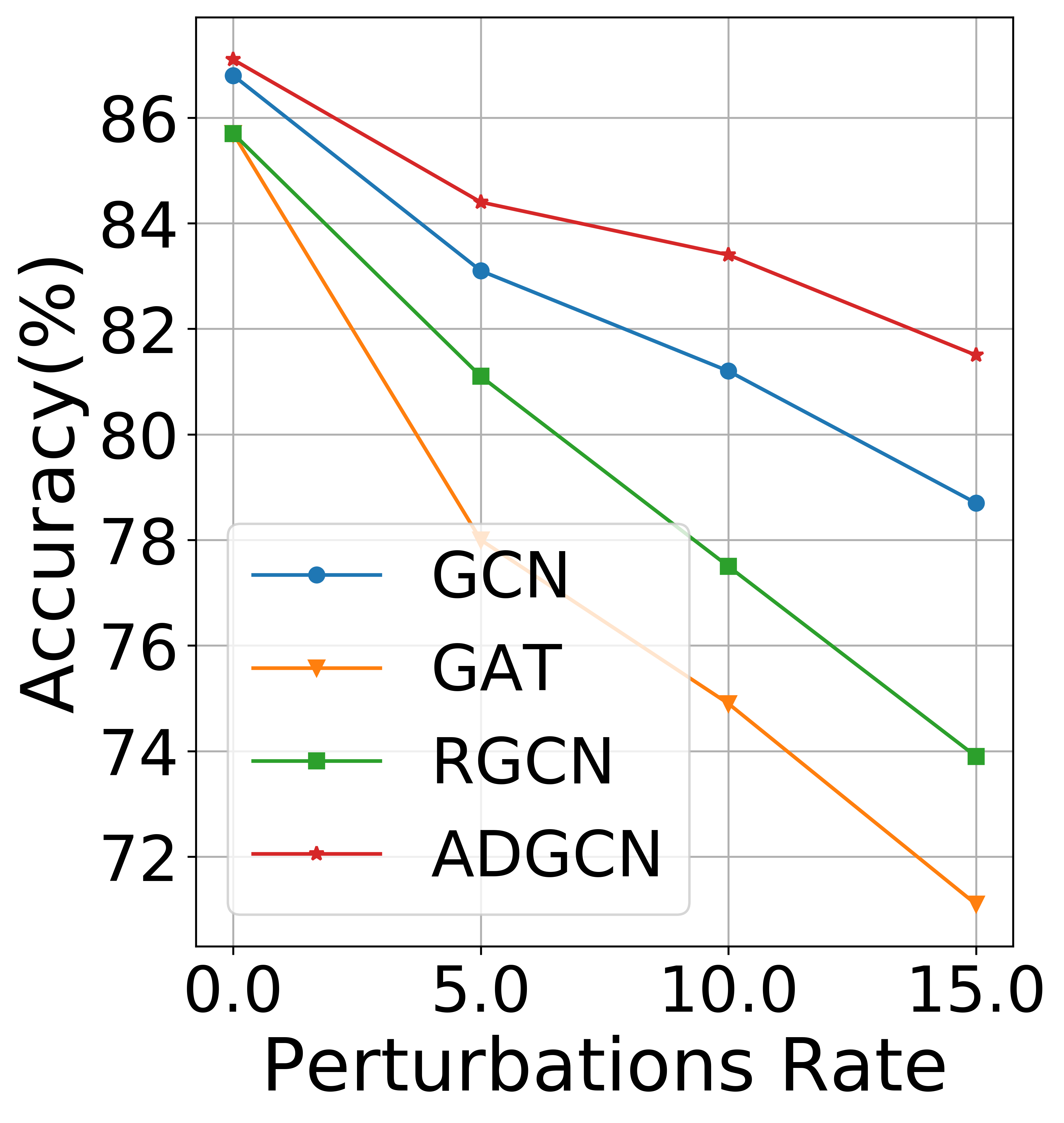}
		%\caption{fig1}
	}%
	\vspace{-0.cm}
	\caption{Node classification performance (\%) of each method on three datasets under two attacks.}
	\label{nettack}
	\vspace{-0.45cm}
\end{figure*}

\textcolor{black}{
% \subsubsection{Efficiency analysis}
\textbf{Efficiency analysis.}
We evaluate the efficiency of the proposed ADGCN from the perspective of computational cost that consists of the GPU memory consumption and time consumption, respectively. }
%They are reported in Tables~\ref{cost},~\ref{each_part_cost}, and~\ref{parameters}.}
\textcolor{black}{As reported in Table~\ref{cost}, the costs in  GPU memory consumption of DisenGCN and ADGCN are almost the same, while the cost is much higher for DisGNN. Particularly, ADGCN-R has the same GPU memory cost as ADGCN because no additional memory consumption is introduced in the graph refinement. }
% \begin{table}[t]
% \centering
% \setlength\tabcolsep{1.3pt}
% \caption{Parameter statistics on six datasets ($\times 10^3$).}
% \begin{tabular}{ccccccc}
% \hline
% Dataset   & GCN   & GAT   & APPNP & DisenGCN & ADGCN & ADGCN-R\\ \hline
% Cora      & 23.1  & 184.7 & 184.7 & 115.3    & 116.8 & 116.8\\
% Citeseer  & 59.4  & 475.2 & 237.4 & 779.1    & 561.7 & 561.7\\
% Pubmed    & 8.1   & 64.7  & 32.2  & 77.6     & 75.1  & 75.1 \\
% CS        & 109.1 & 873.4 & 436.5 & 1023.2   & 684.5 & 684.5\\
% Computers & 12.4  & 99.9  & 49.7  & 108.9    & 105.1 & 105.1\\
% Photo     & 12.1  & 96.8  & 48.2  & 60.3     & 54.0  & 54.0 \\ \hline
% \end{tabular}
% \label{parameters}
% \end{table}
\textcolor{black}{In terms of time consumption from Table~\ref{cost} and ~\ref{each_part_cost}, DisenGCN, ADGCN, and ADGCN-R are not as efficient as FactorGCN due to the involvement of the iterative micro disentanglement. 
%Besides, DisGNN is also the method with the highest time consumption. 
In addition, for both DisenGCN and ADGCN, the time costs are comparable since only a little additional time cost with O$(m)$ will be brought by the macro-D module. What's more, due to the graph refinement process, ADGCN-R is somewhat slower than ADGCN although it achieves better performance.}

% \textcolor{black}{For parameter analysis, Table~\ref{parameters} shows that the number of parameters is comparable between ADGCN and the other competitors except for GCN. Besides, the number of parameters of ADGCN-R is the same as ADGCN since no additional parameters were introduced for the graph refinement.}

\textcolor{black}{
\textbf{Convergence analysis}
% \subsubsection{Convergence analysis}
%Fig.~\ref{macro_disentanglement} has shown that the proposed component-specific aggregation via the dynamic assignment mechanism achieves superior disentanglement performance compared to DisenConv in~\cite{DisenGCN}, while 
%Another issue is whether the dynamic assignment mechanism can converge as well as DisenConv~\cite{DisenGCN}. 
To get a clear view of the convergence of the dynamic assignment mechanism, we define the average difference at iteration $\tilde{t}$ as $b(\tilde{t})=1 - \frac{1}{|V|}\sum_{u \in V}\cos(\mathbf{h}_{u}^{k, (\tilde{t}-1)}, \mathbf{h}_{u}^{k, (\tilde{t})})$, where $\mathbf{h}_{u}^{k, (\tilde{t})}$ denotes the $k$-th component representation of node $u$ at iteration $\tilde{t}$. As shown in Fig.~\ref{convergence}, $b(\tilde{t})$ drops rapidly to a small constant only with a few iterations, which experimentally demonstrates that the convergent of our proposed dynamic assignment mechanism can be guaranteed. \textcolor{revise_color}{Besides, for all 5 components, it can be seen that the curves of ADGCN do decrease faster than the curves of ADGCN w/o $L_{adv}$ on both the Cora and Citeseer datasets. On the Cora dataset, ADGCN is converged at the 3rd iteration while ADGCN w/o $L_{adv}$ is converged at the 4th iteration. Besides, on the Citeseer dataset, ADGCN w/o $L_{adv}$ is slower than ADGCN by 2 iterations to converge.}
% Furthermore, we also provide the loss curves in Fig.~\ref{loss_convergence} and find that the losses of ADGCN w/o Micro-D and ADGCN are gradually decreasing and converging.
}

\vspace{-0.3cm}
\subsection{Defense Performance}
\textcolor{black}{To verify the robustness of ADGCN, we evaluate the node classification performance of ADGCN against two types of classical attacks, i.e., non-targeted attack and targeted attack. Specifically, two typical attacking methods, nettack~\cite{nettack} and metattack~\cite{metattack} are adopted for the targeted attack method and non-targeted attack method, respectively. We first use the attack method to poison the graph, and then we train ADGCN on the poisoned graph and evaluate the node classification performance. It can be seen from Fig.~\ref{nettack} that ADGCN outperforms the three baselines whether under metatack or nettack. In particular, the performance of the baselines varies greatly under different types of attacks, while the performance of ADGCN degrades less than other methods as the attack intensity increases under both attack methods.} 
%These results illustrated the considerable robustness of ADGCN under typical attacking methods compared to the others.}

%\begin{table}[]
%	\centering
%	\caption{Effects evaluation of $\mathcal{L}_{adv}$ and $\mathcal{L}_{adv}$.}
%	\begin{tabular}{l|c|ccc}
%		\hline
%		\multirow{2}{*}{Methods} & \multicolumn{1}{c|}{Classificaiton} & \multicolumn{3}{c}{Clustering} \\ \cline{2-5}
%		& ACC          & ACC          & NMI          & ARI      \\ \hline
%		w/o $\mathcal{L}_{adv}$  & 0.829        & 0.797        & 0.613        & 0.615        \\
%		w/o $\mathcal{L}_{cls}$  & 0.833        & 0.805        & 0.611        & 0.612        \\ \hline
%		ADGCN                    & 0.846        & 0.810        & 0.623        & 0.621        \\ \hline
%	\end{tabular}
%	\label{ablation study}
%	
%\end{table}

\label{HA}
\begin{figure}[t]	
	\centering
	\includegraphics[width=2.3in]{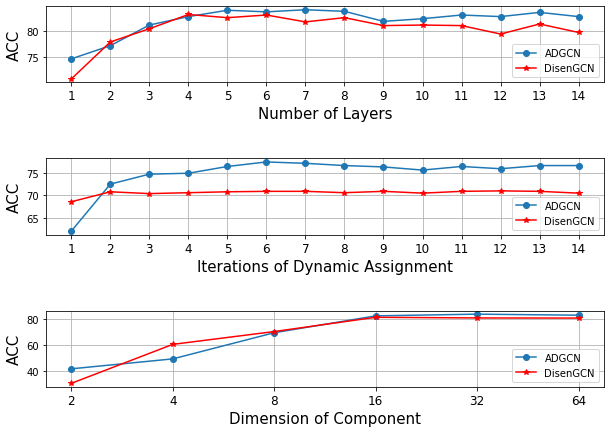}
	\vspace{-0.2cm}
	\caption{Hyper-parameter sensitivity on Cora dataset with the fixed split.}
	\label{hyper}
	%\vspace{-0.2cm}
\end{figure}

\begin{figure}[t]
\vspace{-0.5cm}
	\centering
	\subfigure[($\lambda, \eta$)]{
		\centering
		\includegraphics[width=1.3in]{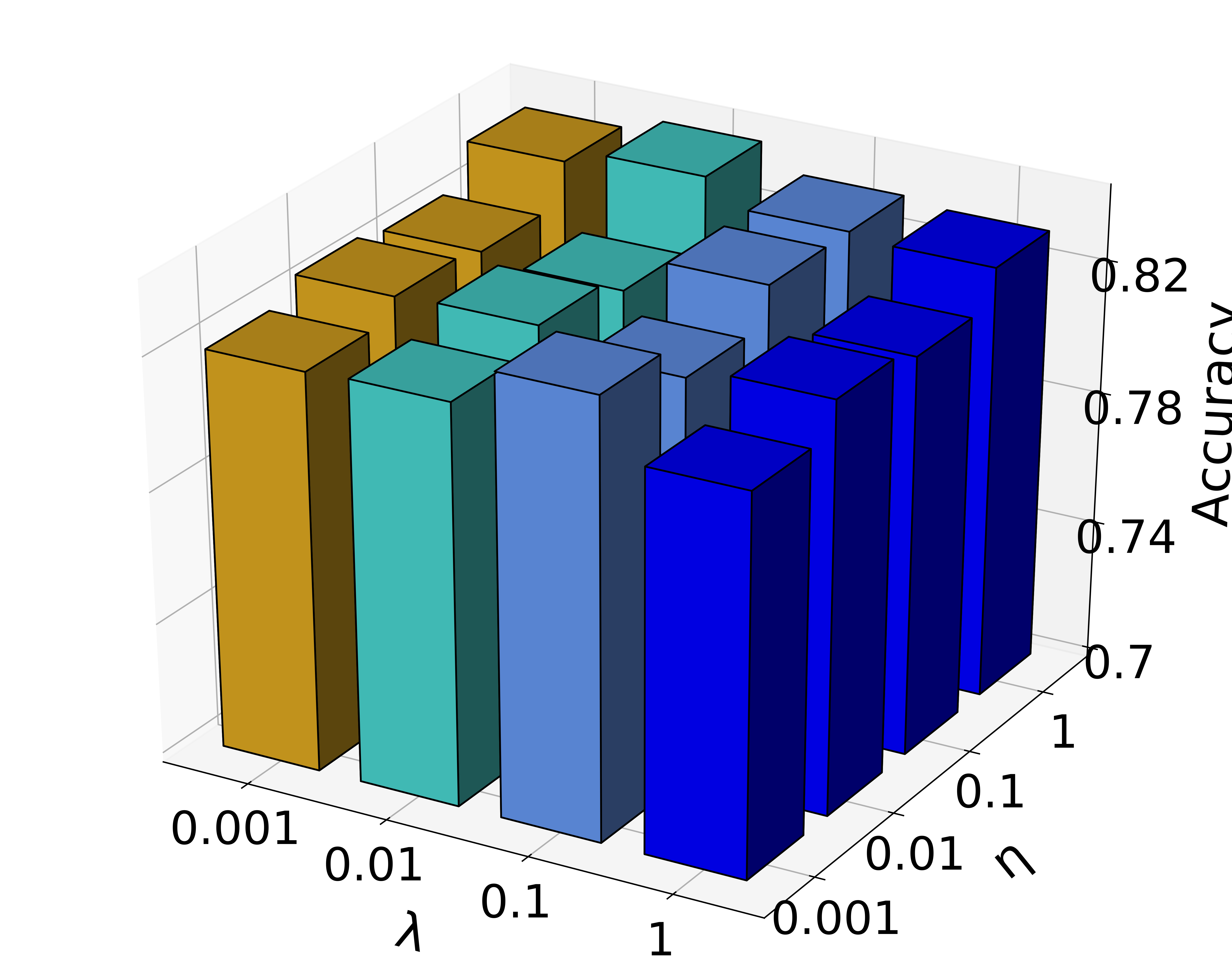}
		%\caption{fig1}
	}%
	\subfigure[($m, \gamma$)]{
		\centering
		\includegraphics[width=1.3in]{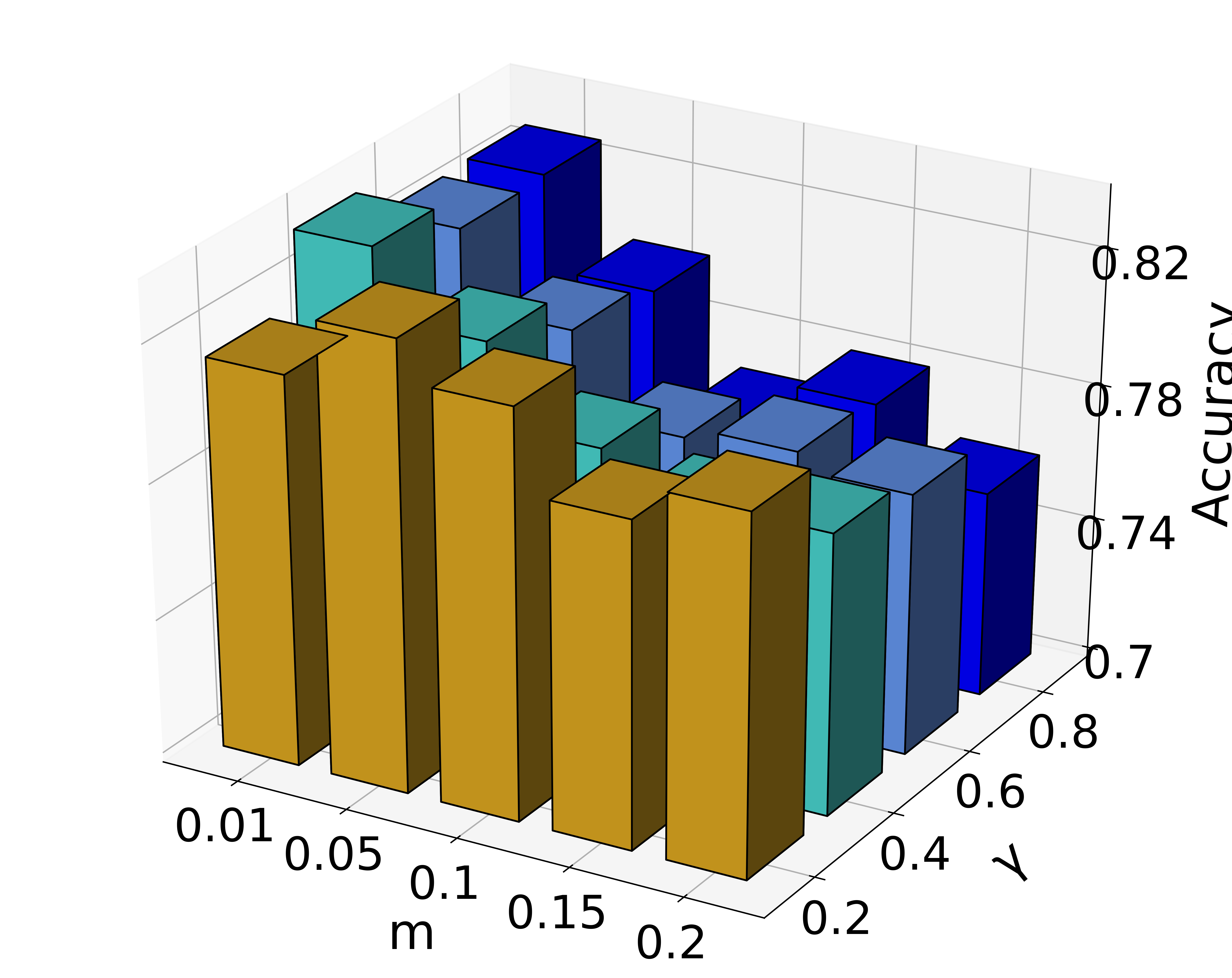}
		%\caption{fig1}
	}%
	\caption{The influences of ($\lambda, \eta$) and ($m, \gamma$) for semi-supervised node classification on Cora dataset.}
	\label{hyper-2}
 \vspace{-0.4cm}
\end{figure}

\subsection{Hyper-parameter Sensitivity Analysis}
We finally study the effect of several hyper-parameters, including the number of layers, the iterations of dynamic assignment $\tilde{T}$, and the dimension of the component, on the Cora dataset with the fixed split.
The results in Fig.\ref{hyper} demonstrate that: (i) with the number of layers increasing, the performance of our model tends to be stable without over-smoothing; (ii) due to the application of the dynamic assignment mechanism, ADGCN can efficiently complete the aggregation in the component space and achieve better performance compared to DisenGCN. \textcolor{black}{Besides, we show the influence of the adversarial loss coefficients ($\lambda$, $\eta$) and the graph refining coefficients ($m$, $\gamma$) in Fig.~\ref{hyper-2}, respectively. Clearly, setting both $\lambda$ and $\eta$ to 1 is a trade-off with good performance on three datasets. Meanwhile, refining the relations with too many nodes sampled from the original graph may reduce the performance of the model, which suggests that setting an appropriate number of nodes to sample is essential for graph refinement.}

\section{Conclusion and Discussion}
In this paper, we propose an ADGCN model that takes both macro-disentanglement and micro-disentanglement into account for graph representation learning.
By micro-disentanglement, it achieves component-specific aggregation of inter-component and intra-component through the dynamic assignment of node neighborhood.
Furthermore, we innovatively introduce the macro-disentanglement adversarial regularizer to explicitly constrain the interdependence between components.
Meanwhile, we also propose a local structure-aware graph refinement based on diversity-preserving node sampling. With the training going on, the graph structure can be progressively refined to reveal the latent relations of nodes.

\textcolor{black}{Although we attempt to carry out a balance of efficiency and effectiveness in graph disentanglement, the proposed ADGCN still suffers from a slight limitation in efficiency caused by the graph refinement and the iterative micro-disentanglement. 
To further reduce time consumption, a faster graph refinement strategy, such as the integration of graph sampling and refinement, is an interesting topic worth exploring.
%In addition, developing a new iterative-free graph disentanglement approach may be another very interesting and challenging work. 
% Based on these possible attempts, our future work will focus on the interpretability of graph disentanglement and apply it to more realistic graph generation.
}

{
	\bibliographystyle{IEEEtran}
	\bibliography{IEEEexample}
}

\end{document}